\documentclass[lettersize,journal]{IEEEtran}
\usepackage{textcomp}
\usepackage{verbatim}
\usepackage{listings}
\usepackage{xspace}
\usepackage{marvosym}

\usepackage{amsmath, amsfonts, amssymb}

\usepackage[table,xcdraw]{xcolor}  
\usepackage{array}
\usepackage{multirow}
\usepackage{arydshln}
\usepackage{threeparttable}
\usepackage{tablefootnote}

\usepackage{graphicx}

\usepackage{algorithm}
\usepackage{algorithmic}

\usepackage{pifont}
\usepackage{bbm}

\usepackage{float}
\usepackage{stfloats}

\usepackage[colorlinks, linkcolor=red, anchorcolor=blue, citecolor=purple, CJKbookmarks=True]{hyperref}

\usepackage{cite}

\definecolor{green}{rgb}{1,0,0}

\usepackage{tcolorbox}

\definecolor{green}{rgb}{1,0,0}

\def\ie{\emph{i.e.}}

\begin{document}

\title{Retrieval-Enhanced Visual Prompt Learning for Few-shot Classification}


\author{Jintao Rong, 
Hao Chen\textsuperscript{\Letter}, 
Linlin Ou, 
Tianxiao Chen, 
Xinyi Yu\textsuperscript{\Letter}, 
Yifan Liu
\thanks{
This work was supported by the National Key R\&D Program of China (No. 2022ZD0118700), the Baima Lake Laboratory Joint Funds of the Zhejiang Provincial Natural Science Foundation of China (Grant No. LBMHD24F030002), and the National Natural Science Foundation of China (Grant No. 62373329).
}
\thanks{Jintao Rong, Linlin Ou, Tianxiao Chen, and Xinyi Yu are with the College of Information Engineering, Zhejiang University of Technology, Hangzhou, China. Email: {\tt\small yuxy@zjut.edu.cn, 202006010503@zjut.edu.cn, 2111903071@zjut.edu.cn, linlinou@zjut.edu.cn}.}
\thanks{Hao Chen is with the College of Computer Science and Technology, Zhejiang University, Hangzhou, China. Email: {\tt\small haochen.cad@zju.edu.cn}.}
\thanks{Yifan Liu is with the School of Computer Science, The University of Adelaide, Adelaide, Australia. Email: {\tt\small yifan.liu04@adelaide.edu.au}.}
\thanks{\textsuperscript{\Letter}Corresponding authors: Hao Chen and Xinyi Yu.}
}



\maketitle

\begin{abstract}
The Contrastive Language-Image Pretraining (CLIP) model has been widely used in various downstream vision tasks. 
The few-shot learning paradigm has been widely adopted to augment its capacity for these tasks.
However, current paradigms may struggle with fine-grained classification, such as satellite image recognition, due to widening domain gaps.
To address this limitation, we propose retrieval-enhanced visual prompt learning (RePrompt), which introduces retrieval mechanisms to cache and reuse the knowledge of downstream tasks.
RePrompt constructs a retrieval database from either training examples or external data if available, and uses a retrieval mechanism to enhance multiple stages of a simple prompt learning baseline, thus narrowing the domain gap.
During inference, our enhanced model can reference similar samples brought by retrieval to make more accurate predictions. A detailed analysis reveals that retrieval helps to improve the distribution of late features, thus, improving generalization for downstream tasks.
Reprompt attains state-of-the-art performance on a wide range of vision datasets, including 11 image datasets, 3 video datasets, 1 multi-view dataset, and 4 domain generalization benchmarks.
\end{abstract}


\section{Introduction}
\IEEEPARstart{V}{isual} concept recognition has achieved remarkable success in closed-set scenarios with large-scale training sets, typically ImageNet dataset. 
The recognition accuracy can even surpass human ability. 
However, it may not always be feasible to have a large training set for each specific visual concept in certain downstream tasks. 
How to learn a robust visual concept recognition system with low-shot data has become a challenging but valuable problem in computer vision. 
Previous methods~\cite{MAML, prototypical, matching} mainly focus on learning a transferable visual representation from a source domain and quickly adapting to few-shot downstream tasks through fine-tuning techniques. 
However, limited by the categories and closed-set training samples in the source domain, these few-shot learning algorithms are only effective in impractically simple settings, such as distinguishing $5$-way $1$-shot classification.

Recent advancements in visual representation learning have been propelled by the emergence of large-scale vision-language models, such as CLIP~\cite{CLIP} and ALIGN~\cite{ALIGN}.
The feasibility of applying these visual language models in addressing more challenging low-shot learning problems has attracted considerable attention within the research community~\cite{CLIP, CoOP, clip_adapter, video_prompt}.
In practice, it is cumbersome to fine-tune the entire vision-language model for transfer learning in each downstream task. 
Furthermore, finetuning the full model can result in catastrophic forgetting~\cite{catastrophic_forgetting,example_forgetting}, leading to poor performance on new tasks despite the pre-trained model's initial success.
To address these challenges, CoOP~\cite{CoOP} first proposes a few-shot evaluation framework for visual language models. 
There are full $C$ classes of downstream tasks and each class has 1/2/4/8/16-shot samples. 
Meanwhile, CoOP designs a learnable text prompt to replace the sub-optimal hand-crafted text prompt templates. 

Nevertheless, the frozen image feature representation also leads to sub-optimal performance. Inspired by VPT~\cite{VPT}, many studies~\cite{UPT, MaPLe, CAVPT, video_prompt, promptsrc} consider adding additional learnable visual tokens into the image encoder to parameter-efficient fine-tune the vision-language models on downstream tasks. 
Despite significant improvements achieved by these methods for few-shot learning, 
these parametric models struggle to generalize to extremely low-shot data or atypical instances, often relying on rote memorization. 
For instance, given only one image per class, CoOP performs even worse than the zero-shot classification results of CLIP.

Retrieval-augmentation approaches~\cite{retrieval_long_tail,RETRO_tokens,RETROPROMPT, retrieval_pretraining, Re-ViLM, tip-adapter} are employed to retrieve knowledge corpus and generate additional references, thereby enhancing performance in low-resource scenarios. 
Large Language Models (LLM) also get advancements with retrieval-augmentation approaches and further develop Retrieval Augmentation Generation (RAG)~\cite{RAG}.
However, the application of retrieval augmentation to visual language models remains underexplored.
Tip-Adapter~\cite{tip-adapter}, a key-value cache model, is constructed as a retrieval auxiliary classifier for CLIP. It leverages the similarity between the test and few-shot training samples, calculating affinity matrixes in text feature space and image feature space.
However, it is observed that the variances and means of visual modal similarities are smaller compared to those in visual-language modalities~\cite{udandarao2022sus}. This phenomenon is caused by CLIP maximizing the cosine similarity between paired samples in different modalities during its contrastive learning training, without considering the intra-modality similarity. Consequently, the image similarity metrics employed by Tip-Adapter, derived from CLIP, are potentially compromised by domain gaps and an insufficient adaptation to downstream datasets.

In this work, we investigate the feasibility of learning prompts from a cache model, which facilitates the enhancement of query image features to adjust image-image similarities. 
To simultaneously optimize these aspects, we introduce Retrieval-enhanced Visual Prompt learning (RePrompt), which incorporates a retrieval mechanism for learning knowledge representations from downstream tasks.
RePrompt establishes a retrieval database that stores common training image representations as retrieval keys. 
The current image representations retrieve relevant cached knowledge from the database.
Unlike Sus-X~\cite{udandarao2022sus}, which expands the retrieval database using external datasets like LAION~\cite{schuhmann2022laion}, our method adheres strictly to the principles of few-shot learning by not utilizing external datasets. Nonetheless, our method can still benefit from incorporating external knowledge.

To integrate the retrieved knowledge into prompt learning, we consider two enhanced strategies based on the nature of the retrieved value. If the retrieved value is the common training image representation, RePrompt dynamically learns retrieval-enhanced visual prompts based on the retrieved values and inserts them into the inputs of multiple layers of the image encoder. 
If the retrieved value corresponds to a label, a learnable $k$NN classifier is applied to predict classification results. These results are linearly interpolated into the final prediction. To harmonize these strategies, we utilize a non-parametric $k$NN algorithm between the query instance and the database. This allows us to establish a prior distribution and construct a loss function that guides the training process.

Our experiments show that RePrompt outperforms previous methods and achieves state-of-the-art performance under various few-shot settings for 11 image datasets, 3 video datasets, and a multi-view dataset. 
Moreover, superior performance on 4 domain generalization benchmarks also demonstrates the generalization ability of our RePrompt for unseen domains.

In summary, the main contributions of our work include:
    \begin{itemize}
    \item{We propose the design of visual Prompt learning with retrieval enhancement, called RePrompt. Specifically, we establish a retrieval database using training examples and implement retrieval-enhanced mechanisms throughout the model, including the input, middle layers, output, and training process.}
    
    \item {We explore the feasibility of introducing a retrieval system to augment the prompt learning for vision language models, which dynamically select relevant references during inference conditioned on the input. This strategy significantly increases the model performance on downstream few-shot classification} 
    
    \item {The proposed RePrompt has strong flexibility and can be easily extended to tasks other than image datasets. For instance, We extend the framework to video understanding and multi-view recognition tasks.}
    
    \item {The proposed RePrompt achieves state-of-the-art performance over 11 image datasets, 3 video datasets, and 1 multi-view image dataset, under various few-shot settings. It also demonstrates superior performance on 4 domain generalization benchmarks.}
    
    \item{We further summarize the patterns between the properties of external memory and retrieval enhancement mechanisms, which will contribute to a better understanding of retrieval enhancement methods in the academic community.}
    \end{itemize}

\section{Related Work}

\noindent\textbf{Retrieval-augmented models} have gained significant attention in the fields of Natural Language Processing (NLP) and Computer Vision (CV). 
Numerous studies have augmented large language models with external memory through retrieval-enhancement mechanism~\cite{dense_retrieval, retrieval_pretraining, bert_knn, RETRO_tokens, RETROPROMPT, retrieval_generation}. 
The integration of knowledge into a language model involves retrieving relevant samples from external memory, thereby enabling the model to generate more informed predictions based on these samples.
Typically, the external memory comprises a collection of text paragraphs or a structured knowledge base.
The BERT-kNN model~\cite{bert_knn} combines the outputs of a trained BERT model with a non-parametric $k$NN model for question-answering tasks. 
REALM~\cite{retrieval_pretraining} acquires external knowledge from Wikipedia. 
Karpukhin et al.~\cite{dense_retrieval} developed a dense passage retrieval system for open-domain question answering. 
RETRO~\cite{RETRO_tokens} demonstrates how to blend retrieved tokens with input tokens in a transformer architecture and methodically evaluates the influence of large-scale external memory datasets on NLP tasks.
Chen et al.~\cite{RETROPROMPT} utilize retrieved tokens as input prompts and fine-tune the entire language model.
Similar to the approach in~\cite{RETROPROMPT}, we employ retrieval samples as prompts to introduce pertinent information.

Recent advancements in CV also leverage external memory for various tasks. 
Several semi-parametric methods~\cite{revisiting_Knn, generalization_memory, tip-adapter, udandarao2022sus, zhang2023prompt}, including Tip-Adapter~\cite{tip-adapter}, have investigated the utilization of a $k$-nearest neighbor (kNN) classifier~\cite{knn_classifier} to enhance classification performance without the need for fine-tuning.
RAC~\cite{retrieval_long_tail} integrates the output of a \emph{base} model, \ie, a conventional vision encoder, with a retrieval module to address long-tail image classification challenges.
Blattmann et al.~\cite{retrieval_diffusion} employ a semi-parametric diffusion model for generative image synthesis, which is augmented by a retrieval module.

Building upon the achievements of retrieval augmentation, we introduce a framework for few-shot classification and domain generalization through retrieval-based techniques. 
Our work focuses on retrieving examples from few-shot training data to construct more dynamic prompts for CLIP models.
The most similar method to our own is \emph{Training-free Adaption} (Tip-Adapter)~\cite{tip-adapter}.
Tip-Adapter solely employs retrieval to enhance the final classification distribution.
We explore the potential of using retrieval to improve the features extracted by CLIP models.

\noindent\textbf{Finetuning for visual language models} is crucial for bridging the domain gap in downstream tasks. 
While CLIP can perform zero-shot image classification by learning reasonable similarity in high-dimensional joint feature space between image-text pairs, 
finetuning with few-shot samples in a parameter-efficient manner remains meaningful. 
To achieve this, CoOP~\cite{CoOP} substitutes the input context token in the text branch of CLIP with learnable parameters~\cite{prefix} for few-shot learning. 
Following recent advancements~\cite{ProDA, ProGrad, prompt_pretraining}, we aim to optimize text prompts more effectively from a gradient perspective. 
CoCoOP~\cite{CoCoOP} proposes to train an intermediate network to generate image tokens as conditional inputs for the text branch and design a base-to-new benchmark to evaluate the
generalization ability of a method within a dataset.
Some studies~\cite{CAVPT, MaPLe, promptsrc, UPT, video_prompt} introduce trainable visual prompts\cite{VPT} into the vision branch of CLIP to develop the dual prompt tuning scheme. 
Maple~\cite{MaPLe, Dualprompt} proposes to learn layer-wise prompts for the two branches simultaneously.
PromptSRC~\cite{promptsrc} utilizes its original features to regularize the learning of prompts.
Beyond their focus on few-shot classification tasks, learnable prompts exhibit remarkable adaptability across diverse downstream applications, offering a unified and efficient framework for fine-tuning foundation models to address various complex tasks~\cite{lu2024prompt, Jin2024Domain, Zeng2024Temporally, zhu2024Prompt, huang2024Cross}.

Adapters are typically subnetworks consisting of two fully connected layers with a nonlinear activation function in between. 
Referring to the concept of adapters~\cite{parameter-efficient}, CLIP-Adapter~\cite{clip_adapter} employs lightweight adapters to adapt final features. 
Tip-Adapter`\cite{tip-adapter} utilizes a key-value cache model for auxiliary classification.
DPT~\cite{Dualprompt} propose class-aware visual prompts to guide the CLIP model focus on the target object in downstream work.
Sus-X~\cite{udandarao2022sus} constructs a support set using generative models and extends Tip-Adapter by leveraging image-text affinity matrices. 
CaFO~\cite{zhang2023prompt} improves visual recognition in low-shot regimes by expanding the training set through a generative model and integrating prior knowledge from multiple pre-training models using multiple adapters.

RePromp provides an in-depth analysis of the key-value cache model from a retrieval perspective. It further explores the potential of this model to contribute to CLIP fine-tuning through retrieval-enhanced visual prompt learning. 
This approach enables RePrompt to effectively adjust the distribution of the image-feature space utilized in Tip-Adapter.

\section{Preliminaries}\label{sec:preliminaries}
CLIP~\cite{CLIP} is a pre-trained model that learns aligned vision-language representations from web-scale image-text datasets. 
It comprises two sub-networks: an image encoder $\mathrm{e}_\mathrm{I}$ and a text encoder $\mathrm{e}_\mathrm{T}$. 
The image encoder encodes visual inputs, and the text encoder encodes text inputs, into a shared hidden space $\mathbb{R}^d$. 
Here, $d$ represents the dimension of embeddings (e.g.,d = 512 for ViT\cite{ViT} in CLIP).

 \begin{figure}[ht]
  \centering
  \includegraphics[width=1.0\linewidth]{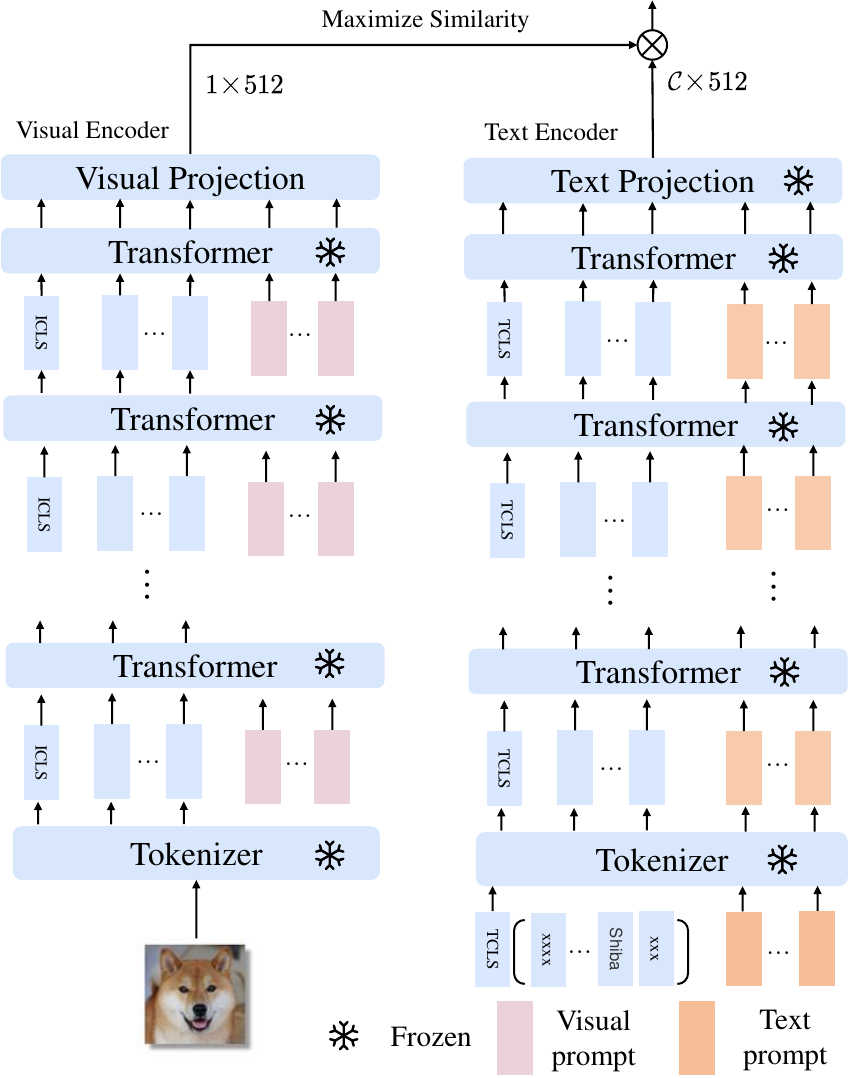}
   \caption{The overall architecture of vision-language prompt tuning. Visual and textual prompt tokens, which are the only learnable parameters in this setup, are incorporated into the vision and language branches of CLIP, respectively. These prompts are designed to optimize performance in low-shot scenarios while preserving the model's original generalization capabilities.}
   \label{fig:VLPT}
   \vspace{-0.5cm}
\end{figure}

\noindent\textbf{Classification of CLIP.}
Given an image $\boldsymbol{x}$ and a set of $C$ category names $T = \{\boldsymbol{t}_1, \boldsymbol{t}_2, \dots, \boldsymbol{t}_C\}$ (e.g., $C=1000$ for ImageNet~\cite{imagenet_paper}), the image encoder extracts the image feature $\boldsymbol{z} = e_{\mathrm{I}}(\boldsymbol{x}) \in \mathbb{R}^d$. 
The category names in $T$ are combined with a hand-crafted text prompt template ``a photo of a [CLASS]" to generate text descriptions.
These descriptions are then fed into the text encoder to derive the text features $\boldsymbol{f} \in \mathbb{R}^{d \times C}$.
The prediction probability of $\boldsymbol{x}$ belonging to class $c$ is calculated using the inner product similarity between the query image feature $\boldsymbol{z}$ and text features $\boldsymbol{f}$ as following: 
\begin{equation}\label{zero_shot_clip}
p(c|x) = \frac{\exp \left( \left( \boldsymbol{z} \cdot \boldsymbol{f}_c \right) / \tau \right)}{\sum_{c=1}^{C} \exp \left( \left( \boldsymbol{z} \cdot \boldsymbol{f}_c \right) / \tau \right)},
\end{equation}
where $\tau$ is the temperature coefficient learned by CLIP. The text encoder produces a retrieval-based dynamic classifier.

\noindent\textbf{Text prompt learning.} 
In contrast to hand-crafted prompt engineering, text prompt learning aims to generate more adaptive text features by learning a set of prompts.
CoOp~\cite{CoOP} learns a set of parameters $\boldsymbol{P}_\mathrm{T} \in \mathbb{R}^{d \times M}$ to replace the predefined prompt templates, where $M$ represents the prompt length.
The word token of each category name in $T$ is inserted into a template and treated as the initial values of learnable input prompts.
These prompts are further fine-tuned on few-shot data to generate text features.
The fine-tuned prompts $\boldsymbol{P}_\mathrm{T}$ adjust the decision boundaries of text features for downstream tasks.
All parameters inherited from the pretrained CLIP model remain fixed during the training process.

\noindent\textbf{Visual prompt learning.}
VPT~\cite{VPT} proposes a visual prompt tuning approach, which introduces a few learnable parameters into the input space while freezing the image encoder $e_{\mathrm{I}}$.
This method aims to extract more transferable visual features from downstream data.
In the context of an image encoder with $L$ layers, the output of the $i$-th layer, $l_i$, for $i=1,2,\dots,L$, can be expressed as:
\begin{equation}\label{vit_output}
\left[ \boldsymbol{c}, \boldsymbol{z}^{i+1}_{1}, \dots, \boldsymbol{z}^{i+1}_{S} \right] = l_{i}\left( \boldsymbol{c}, \boldsymbol{z}^{i}_{1}, \dots, \boldsymbol{z}^{i}_{S} \right),
\end{equation}
where $\boldsymbol{c} \in \mathbb{R}^d$ denotes the classification token and $Z^{i} = \{\boldsymbol{z}^{i}_{1}, \dots, \boldsymbol{z}^{i}_{S}\} \in \mathbb{R}^{d \times S}$ represents a set of input image patch tokens of the $i$-th layer, with token sequence length $S$.
Furthermore, the learnable visual prompt $P_{\mathrm{I}} \in \mathbb{R}^{d \times N}$ is introduced into the input sequence of the $i$-th layer. $N$ is the length of the visual prompt.
There are two visual prompt variants, VPT-Shallow and VPT-Deep. In VPT-Shallow, the class token $\boldsymbol{c}$, along with image patch tokens and visual prompts, is taken as the input of the first layer. VPT-Deep inserts independent visual prompts into each layer.

In this section, we propose a straightforward baseline approach for few-shot image classification, named Vision-Language Prompt Tuning (VLPT). Fig.~\ref{fig:VLPT} illustrates the overall architecture of VLPT.
The visual prompt is optimized through visual prompt learning, while the text prompt is refined via text prompt learning. 
Upon deriving the image and text features for each category, the model with tuned prompts generates predictions based on the classification paradigm of CLIP.

\begin{figure*}[htp]
  \centering
  \includegraphics[width=1.0\linewidth]{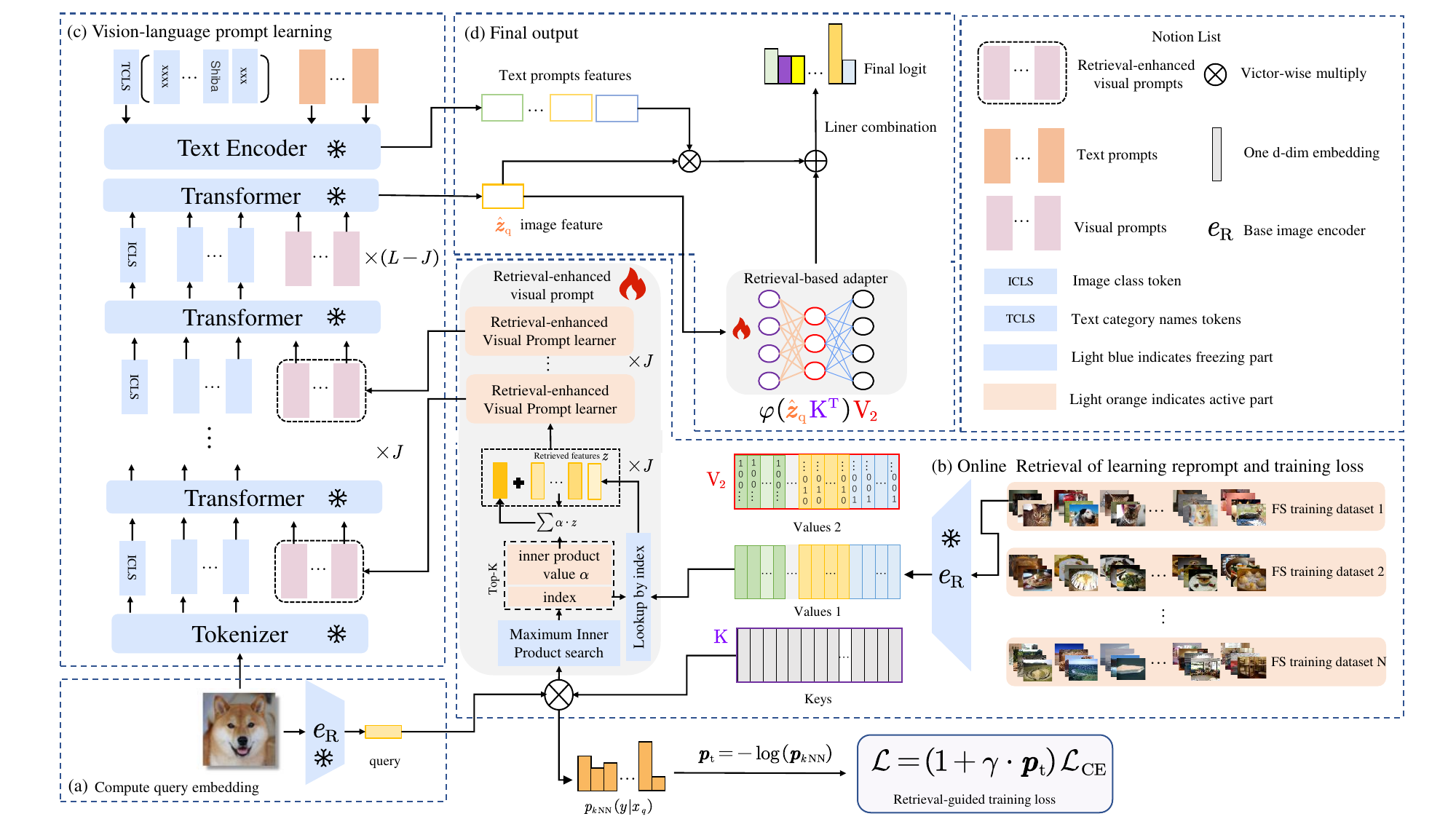}
  \vspace{-5mm}
    \caption{\textbf{The overall workflow of RePrompt} includes four main steps: (a) Encoding an image input into a query embedding using a frozen image encoder; (b) Encoding each image entry from the training dataset into key and value embedding pairs with the same frozen image encoder, where the value embeddings also include one-hot representations of labels. We retrieve the top-K relevant knowledge items through maximum inner product search and integrate this knowledge to generate visual prompts; (c) Retrieval-enhanced visual prompts are introduced into the $J$ layers of the visual branch inputs, while the other prompts remain consistent with those of the baseline VLPT; (d) The final output is derived from a linear combination of the prompt-tuned CLIP prediction and the retrieval-aided prediction. The retrieval database for each dataset comes from a corresponding few-shot training set for fairness.
    }
   \label{fig:RePrompt}
   \vspace{-3mm}
\end{figure*}

\section{Proposed Method}\label{sec:methods}
This section provides a detailed explanation of the proposed Retrieval-enhanced Prompt Tuning for VLPT. RePrompt enhances prompt tuning by utilizing the retrieval of relevant information from the training dataset.
Given an input image $x$, RePrompt retrieves $k$ potentially useful entries from the retrieval database. Each entry is associated with a corresponding representation $z$ and a label (the construction of the retrieval database is described in Section~\ref{subsec:retrbasis}). The retriever uses embedding similarity to find relevant downstream knowledge in the retrieval database. We then condition both the retrieved set and the original input $x$ to generate the output through the retrieval-enhanced visual prompt and retrieval-based adapter. 
Fig.~\ref{fig:RePrompt} illustrates the workflow of RePrompt. Specifically, Section~\ref{subsec:retrbasis} describes the process of retrieving entries most relevant to the input query. Subsequently, Section~\ref{subsec:reprompt} details the prompt learner that integrates the query and retrieved knowledge into the prompt learning process. In Section~\ref{subsec:rbadapter}, we discuss the final prediction in the inference process using the $k$NN-based probability from the retrieval-based adapter. Additionally, this section describes how non-parametric KNN is used between the query and the database to obtain probability distributions, while dynamically controlling the strength of the retrieval mechanism.

\subsection{Retrieval Module}\label{subsec:retrbasis}
Our retrieval mechanism/process consists of two steps: (1) retrieval database construction, and (2) nearest neighbor retrieval.

\noindent\textbf{Retrieval database construction.} The retrieval database is constructed by extracting features from the few-shot training dataset $\mathcal{D}$. 
Specifically, the database contains $\left| \mathcal{D} \right|$ key-value pairs $\left( \boldsymbol{k}_i, \boldsymbol{v}_i \right)$. 
Each key $\boldsymbol{k}_i = \boldsymbol{e}_{\mathrm{R}}(\boldsymbol{x}_i) \in \mathbb{R}^{d}$ represents the training image representation encoded by the frozen image encoder $\boldsymbol{e}_{\mathrm{R}}$. 
In our method, values of each key-value pair comprise two components: the label $\boldsymbol{y} \in \mathbb{N}^+$ (one-hot encoding) and the image representation $\boldsymbol{e}_{\mathrm{R}}(\boldsymbol{x}_i) \in \mathbb{R}^{d}$ encoded by $\boldsymbol{e}_{\mathrm{R}}$. 
The retrieval database acts as a repository of robust tokens that are adaptable to changes in downstream tasks.

Previous studies~\cite{retrieval_long_tail,retrieval_pretraining, RETRO_tokens} have demonstrated the effective performance improvements achieved by expanding the retrieval database. 
Direct use of external data can disrupt the few-shot task setting. 
\textbf{To ensure a fair comparison, synthesized data are not included in the retrieval database used for main experiments.}
In the ablation study,
we employ a stable diffusion generation model~\cite{rombach2022high} to generate additional training data to expand training data without requiring additional human effort for data collection and annotation. 
The synthesized images are added to the original retrieval database.
In Table~\ref{tab:componet}, we present the performance improvement achieved by incorporating synthesized data, demonstrating an average precision of 84.19\% for the 16-shot scenario.

\noindent\textbf{Nearest neighbor retrieval.}
As shown in Fig.~\ref{fig:RePrompt}, the retrieval database is represented as a matrix $\boldsymbol{D} \in \mathbb{R}^{\left| \mathcal{D} \right| \times d}$, which serves as a fast approximate $k$-NN of examples. 
Given a query image $\boldsymbol{x}_{\mathrm{q}}$, we utilize the encoder $e_\mathrm{R}$ to map it to a corresponding vector $\boldsymbol{z}_{\mathrm{q}} = \boldsymbol{e}_{\mathrm{R}}(\boldsymbol{x}_{\mathrm{q}})$. 
Using the query vector $\boldsymbol{z}_{\mathrm{q}}$, we retrieve its approximate $k$-nearest neighbors with representations $\boldsymbol{z}_1, \boldsymbol{z}_2, \dots, \boldsymbol{z}_k$ from the matrix $\boldsymbol{D}$ based on cosine similarity. 
The image encoder of CLIP is proved to be an effective $e_\mathrm{R}$ in identifying more accurately matched images.
To ensure an efficient retrieval process, we employ FAISS~\cite{faiss} for querying the database.


\subsection{Retrieval-enhanced Visual Prompting} \label{subsec:reprompt}
The proposed method aims to enhance visual prompt learning by utilizing a retrieval database for analogy learning. The visual prompts generated from the retrieval results are referred to as retrieval-enhanced visual prompts.

The retrieval module receives a query vector $\boldsymbol{z}_{\mathrm{q}}$ derived from a raw image $\boldsymbol{x}_{\mathrm{q}}$ and performs a lookup operation in the matrix $\boldsymbol{D}$ to obtain the top $k_{\mathrm{re}}$ most similar candidates. The corresponding representations $\boldsymbol{z}_1, \boldsymbol{z}_2, \dots, \boldsymbol{z}_{k_{\mathrm{re}}}$ retrieved from the database are integrated into the image encoder to enhance the visual prompts. 
Additional fusion vectors $\boldsymbol{z}_{\mathrm{f}} \in \mathbb{R}^d$~\cite{RETROPROMPT} are generated by intuitively aggregating $k_{\mathrm{re}}$ neighboring representations based on their similarity, as follows:
\begin{equation}\label{z_f}
\boldsymbol{z}_{\mathrm{f}} = \sum_{i=1}^{k_{\mathrm{re}}} \alpha_i \cdot \boldsymbol{z}_i,\quad \alpha_i = \frac{e^{\boldsymbol{z}_{\mathrm{q}} \cdot \boldsymbol{z}_i}}{\sum_{j=1}^{k_{\mathrm{re}}} e^{\boldsymbol{z}_{\mathrm{q}} \cdot \boldsymbol{z}_j}}.
\end{equation}
The query vectors $\boldsymbol{z}_{\mathrm{q}}$, fusion vectors $\boldsymbol{z}_{\mathrm{f}}$, and retrieved vectors $\boldsymbol{z}_1, \boldsymbol{z}_2, \dots, \boldsymbol{z}_{k_{\mathrm{re}}}$ are concatenated to form the input $\hat{\boldsymbol{i}} = \left[\boldsymbol{z}_{\mathrm{q}}, \boldsymbol{z}_{\mathrm{f}}, \boldsymbol{z}_1, \dots, \boldsymbol{z}_{k_{\mathrm{re}}}\right] \in \mathbb{R}^{d \times (k_{\mathrm{re}} + 2)}$ for the visual prompt learner, which then generates the retrieval-enhanced visual prompts $f_{\mathrm{p}}(\hat{\boldsymbol{i}}) \in \mathbb{R}^{d \times (k_{\mathrm{re}} + 2)}$.

\begin{figure}[h]
  \centering
  \includegraphics[width=0.8\linewidth]{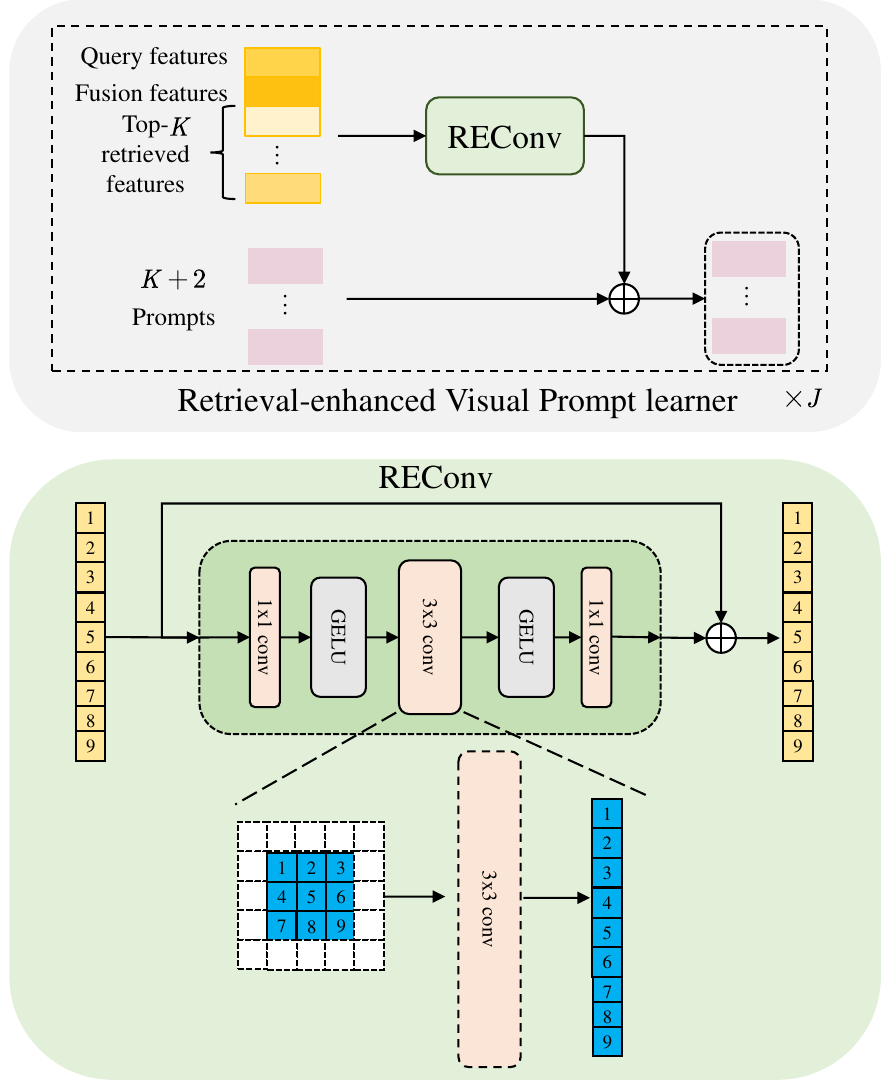}
  \vspace{-3mm}
   \caption{\textbf{Overview of visual prompt learner and REConv}. A visual prompt learner comprises the REConv to generate dynamic visual prompts by learning on retrieved results}
   \label{fig:promptlearner}
   \vspace{-5mm}
\end{figure}

As depicted at the top of Fig.~\ref{fig:promptlearner}, we randomly initialize $J$ visual prompt sets, $\boldsymbol{P}^{1}_{\mathrm{I}}, \dots, \boldsymbol{P}^{J}_{\mathrm{I}}$. Subsequently, $J$ retrieval-enhanced convolution (REConv) blocks process $J$ inputs, $\hat{\boldsymbol{i}}_{1},\dots,\hat{\boldsymbol{i}}_{J}$ (by replicating $\hat{\boldsymbol{i}}$ $J$ times), to generate dynamic prompts. These prompts are combined separately with the $J$ visual prompt sets and inserted into the input sequences of the first $J$ layers of the visual branch:
\begin{equation}
\label{Re_prompt}
\left[\boldsymbol{c},\boldsymbol{Z}^{\boldsymbol{j}+1}\right] = l_{\boldsymbol{j}}\left(\boldsymbol{c}, f_{\mathrm{p}}(\hat{\boldsymbol{i}}_{\boldsymbol{j}}) + \boldsymbol{P}_{\mathrm{I}}^{\boldsymbol{j}}, \boldsymbol{Z}^{\boldsymbol{j}}\right),\boldsymbol{j}=1,\dots ,J.
\end{equation}
The output token corresponding to the visual prompt is discarded. The remaining $12-J$ layers process learnable prompts without enhancement, and learning these prompts reverts to conventional visual prompt tuning. 

We propose a retrieval-enhanced convolution block to effectively and efficiently fuse the retrieved representations (REConv).
As illustrated at the bottom of Fig.~\ref{fig:promptlearner}, a REConv block comprises three convolution layers: two $1\times1$ convolutions, which individually reduce and scale the channel dimensionality and a $3\times3$ convolution positioned between the two $1\times1$ convolutions. Before these convolution layers, we reshape the 1D token sequence structure of visual prompts into a 2D matrix structure.
REConv blocks process the input $\hat{\boldsymbol{i}}$ in parallel to generate dynamic prompts, which can be formulated as
\begin{equation}\label{eq9}
f_{\mathrm{p}}(\hat{\boldsymbol{i}}_{\boldsymbol{j}}) = \beta \mathrm{REconv}_{\boldsymbol{j}}(\mathrm{LN}(\hat{\boldsymbol{i}}_{\boldsymbol{j}})) + \hat{\boldsymbol{i}}_{\boldsymbol{j}}, \quad \boldsymbol{j}=1,\dots,J.
\end{equation}
In this context, $\mathrm{LN}$ denotes Layer Normalization, and $\beta$ is a hyper-parameter used for scaling the output.

\subsection{Retrieval-based Adapter} \label{subsec:rbadapter}
We regard Tip-Adapter as a retrieval-based adapter from the perspective of a $k$NN classifier framework. 
The adapter is trained in conjunction with the retrieval-enhanced visual prompt to enhance the generation of adaptive prediction probabilities for downstream tasks.
The query vector, $\hat{\boldsymbol{z}}_{\mathrm{q}}$, is derived from a query instance, ${\boldsymbol{x}}_{\mathrm{q}}$, using an image encoder that integrates retrieval-enhanced visual prompts. This vector retrieves the $\left| \mathcal{D} \right|$ nearest neighbors and their corresponding inner product similarities.
We aggregate the probability mass for each label, ${\boldsymbol{y}}_i$, across all its occurrences among the retrieved targets. Assuming $\boldsymbol{p}_{k\mathrm{NN}}$ represents the probability of the query instance ${\boldsymbol{x}}_{\mathrm{q}}$ being classified as label $\boldsymbol{y}$, we then reformulate $\boldsymbol{p}_{k\mathrm{NN}}\left( \boldsymbol{y} \mid \boldsymbol{x}_{\mathrm{q}} \right)$ using a weighted sum of the $\boldsymbol{p}_{k\mathrm{NN}}$ probabilities as follows:
\begin{equation}\label{eq10}
\boldsymbol{p}_{k\mathrm{NN}}\left( \boldsymbol{y} \mid \boldsymbol{x}_{\mathrm{q}} \right) \propto \sum_{\left( {\boldsymbol{z}}_i, {\boldsymbol{y}}_i \right) \in \boldsymbol{D}} \mathbf{1}_{ {\boldsymbol{y}}={\boldsymbol{y}}_i} \exp \left( \hat{{\boldsymbol{z}}}_{\mathrm{q}} \cdot {\boldsymbol{z}}_i \right).
\end{equation}
To better integrate the prediction from the $k$NN and the base model, we interpolate $\boldsymbol{p}\left( \boldsymbol{y} \mid \boldsymbol{x}_{\mathrm{q}} \right)$, blending the $k$NN prediction with the CLIP prediction, scaled by a factor $\lambda$:
\begin{equation}\label{eq11}
\boldsymbol{p}\left( \boldsymbol{y} \mid \boldsymbol{x}_{\mathrm{q}} \right) = \lambda \boldsymbol{p}_{k\mathrm{NN}}\left( \boldsymbol{y} \mid \boldsymbol{x}_{\mathrm{q}} \right) + (1-\lambda) \boldsymbol{p}_{\mathrm{P}}\left( \boldsymbol{y} \mid \boldsymbol{x}_{\mathrm{q}} \right).
\end{equation}

\noindent\textbf{Relation to Tip-Adapter.}
Tip-Adapter constructs an adapter using a key-value cache model from the few-shot training set. The cache model stores prior knowledge encoded in CLIP and facilitates image classification through retrieval in the image embedding space. Tip-Adapter is considered as an auxiliary classifier that leverages image-image similarity. While both approaches aim to improve performance using retrieval, our paper highlights two key distinctions:
\textbf{(1)} The features extracted by the CLIP image encoder with retrieval-enhanced prompts are more adaptive to downstream tasks compared to the features obtained by the original CLIP image encoder in Tip-Adapter. As a result, prompt-tuned query features can be utilized to compute improved image-image and image-text similarity matrices for classification.
\textbf{(2)} We extract potential visual context from the cache model and integrate it into the model in the form of prompts, rather than considering the cache model solely as an adapter for classification assistance.

\subsection{Retrieval-guided Training} \label{subsec:rgtraining}
The $k$-nearest neighbor ($k$NN) algorithm primarily focuses on approximating the neighborhoods of query instances\cite{knn_classifier}. It is intuitive to leverage $k$NN classification results as prior knowledge to guide RePrompt's focus on hard examples during the training process. Hard samples usually refer to atypical samples with low confidence. To compute a local probability distribution, we limit the number of samples within the retrieved neighbor set $\boldsymbol{K} \subseteq \boldsymbol{D}$, where $k_{rc} \ne \left| \boldsymbol{K} \right|$, as follows:
\begin{equation}\label{eq12}
    \boldsymbol{p}_{k\mathrm{NN}}\left( \boldsymbol{y} \mid \boldsymbol{x}_{\mathrm{q}} \right) \propto \sum_{\left( {\boldsymbol{z}}_i, {\boldsymbol{y}}_i \right) \in \boldsymbol{K}} \mathbf{1}_{ {\boldsymbol{y}} = {\boldsymbol{y}}_i} \exp \left( {\boldsymbol{z}}_{\mathrm{q}} \cdot {\boldsymbol{k}}_i \right).
\end{equation}
The probability $\boldsymbol{p}_{k\mathrm{NN}}$ represents the confidence of classifying the query instance $\boldsymbol{x}_{\mathrm{q}}$ into specific categories. The negative log-likelihood value of $\boldsymbol{p}_{k\mathrm{NN}}$, similar to Focal Loss\cite{focal_loss}, serves as the adjustment factor $\boldsymbol{p}_{\mathrm{t}} = -\log \left( \boldsymbol{p}_{k\mathrm{NN}} \right)$. This adjustment factor modifies the relative loss of pseudo-correct or pseudo-error samples distinguished by $k$NN, thereby reweighting the cross-entropy loss $\mathcal{L}_{\mathrm{CE}}$. The final loss is formulated as follows:
\begin{equation}\label{eq13}
    \mathcal{L} = \left( 1 + \gamma \cdot \boldsymbol{p}_{\mathrm{t}} \right) \mathcal{L}_{\mathrm{CE}},
\end{equation}
where $\gamma$ is a scaling factor. We set $\left| \boldsymbol{K} \right| = C \times \boldsymbol{n}, \boldsymbol{n} \in \mathbb{N}^{+}$\cite{RETROPROMPT}, which adopts a similar loss for augmenting the model performance on the few-shot learning task of NLP. In few-shot experiments, $\boldsymbol{n}$ can take values of $1, 2, 4, 8, 16$ to accommodate few-shot settings. We discuss the choice of $n$ in Table~\ref{tab:loss_n_more}.
\section{Experiments}

\begin{figure*}
	\centering
	\begin{minipage}{0.32\linewidth}
		\centering
		\includegraphics[width=0.9\linewidth]{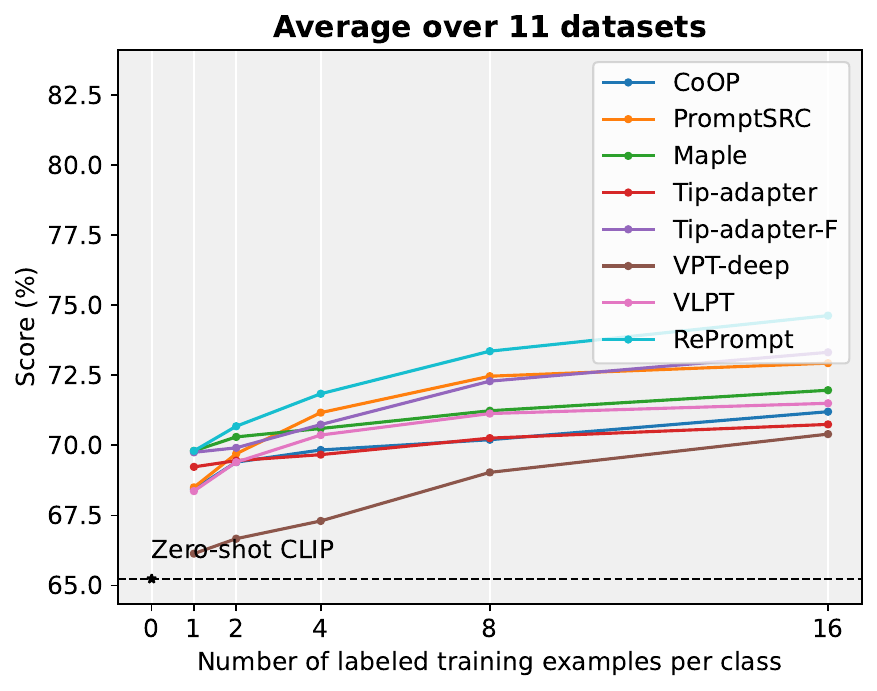}
	\end{minipage}
	\begin{minipage}{0.32\linewidth}
		\centering
		\includegraphics[width=0.9\linewidth]{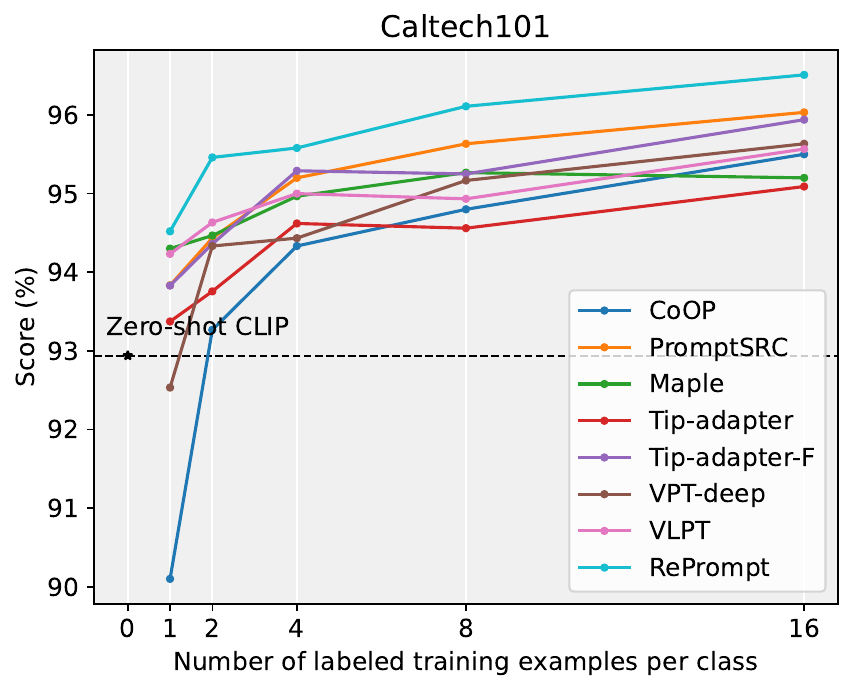}
	\end{minipage}
	\begin{minipage}{0.32\linewidth}
		\centering
		\includegraphics[width=0.9\linewidth]{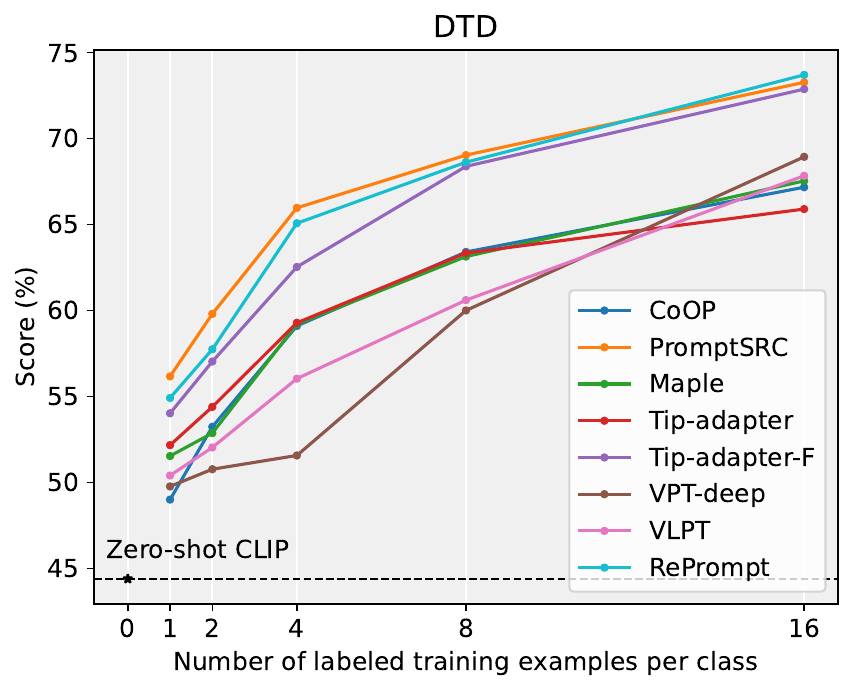}
	\end{minipage}
	\qquad
	\begin{minipage}{0.32\linewidth}
		\centering
		\includegraphics[width=0.9\linewidth]{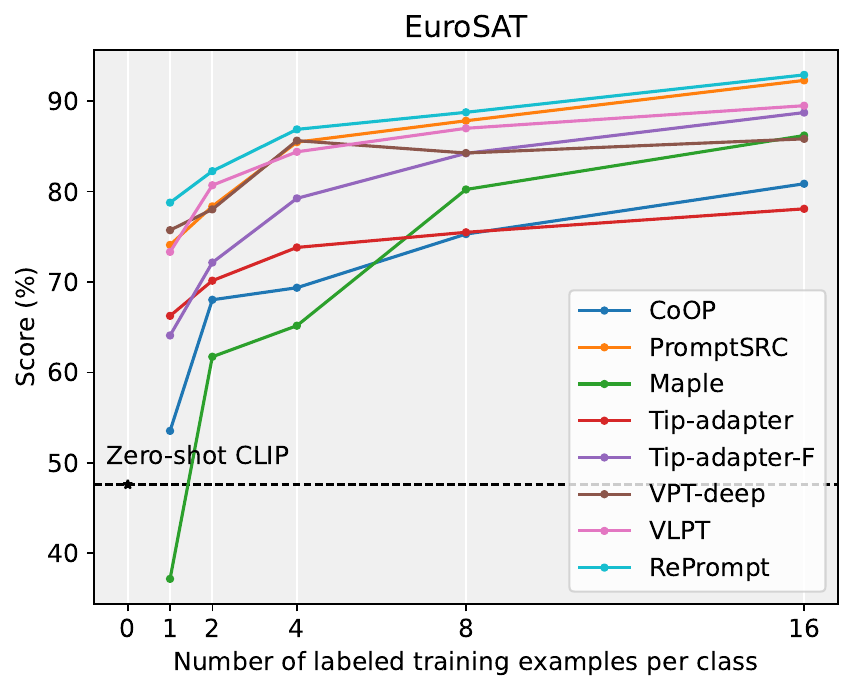}
	\end{minipage}
	\begin{minipage}{0.32\linewidth}
		\centering
		\includegraphics[width=0.9\linewidth]{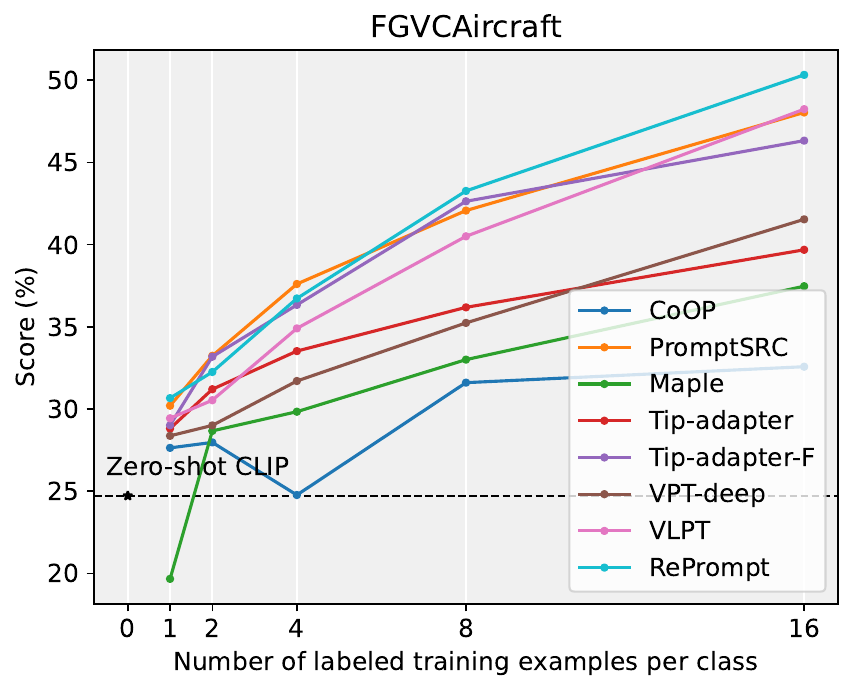}
	\end{minipage}
	\begin{minipage}{0.32\linewidth}
		\centering
		\includegraphics[width=0.9\linewidth]{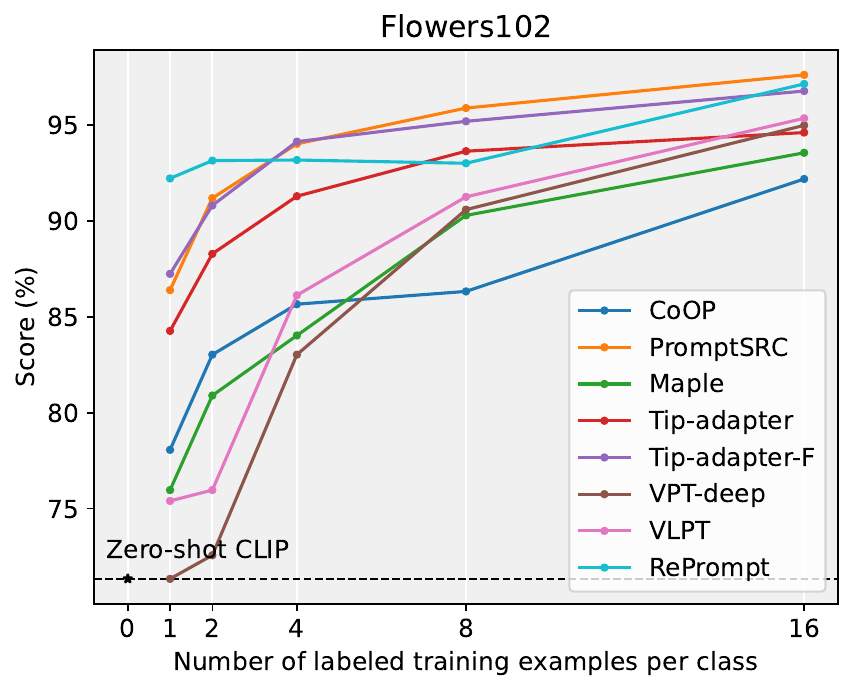}
	\end{minipage}
	\qquad
	\begin{minipage}{0.32\linewidth}
		\centering
		\includegraphics[width=0.9\linewidth]{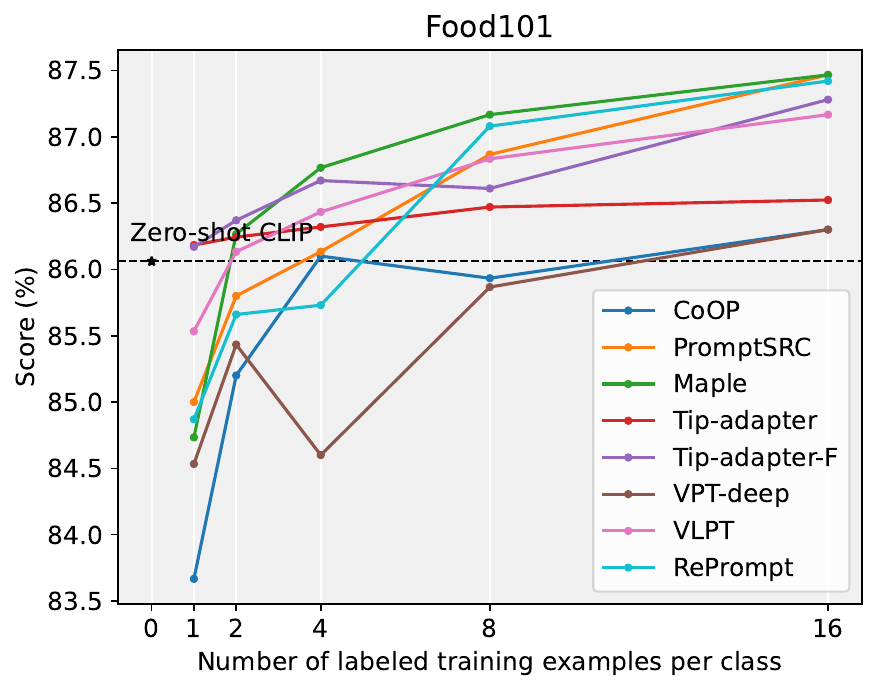}
	\end{minipage}
	\begin{minipage}{0.32\linewidth}
		\centering
		\includegraphics[width=0.9\linewidth]{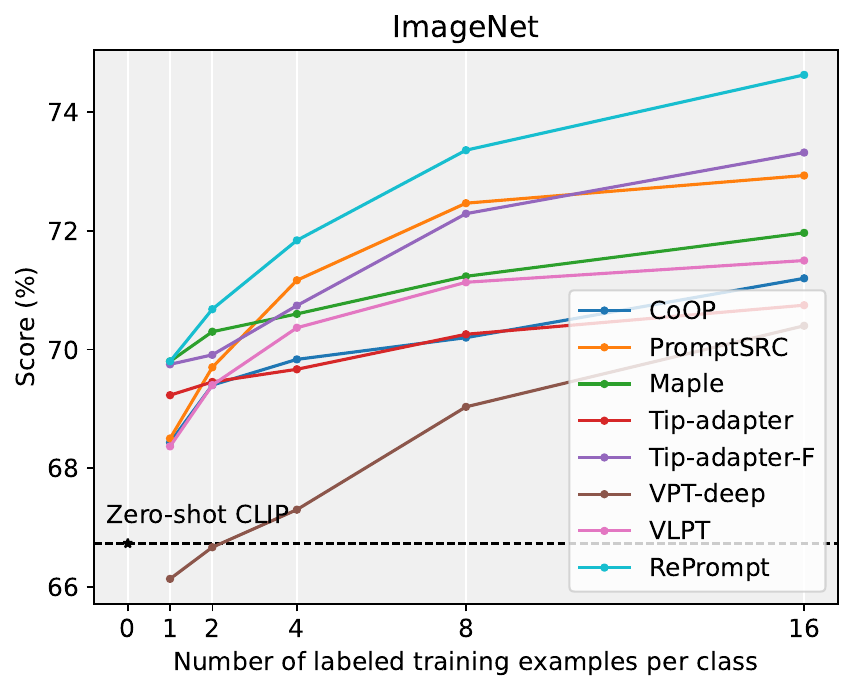}
	\end{minipage}
	\begin{minipage}{0.32\linewidth}
		\centering
		\includegraphics[width=0.9\linewidth]{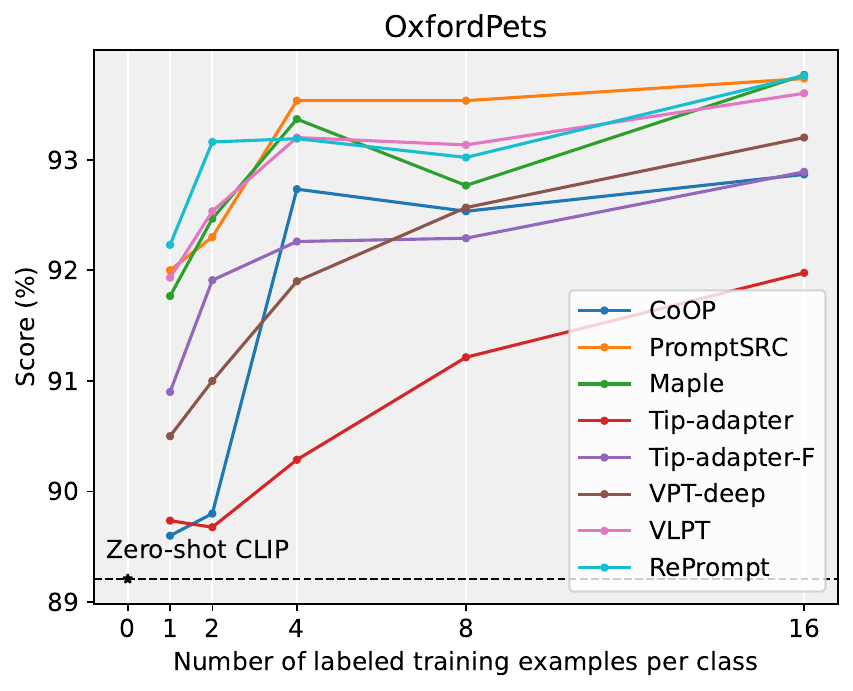}
	\end{minipage}
	\qquad
	\begin{minipage}{0.32\linewidth}
		\centering
		\includegraphics[width=0.9\linewidth]{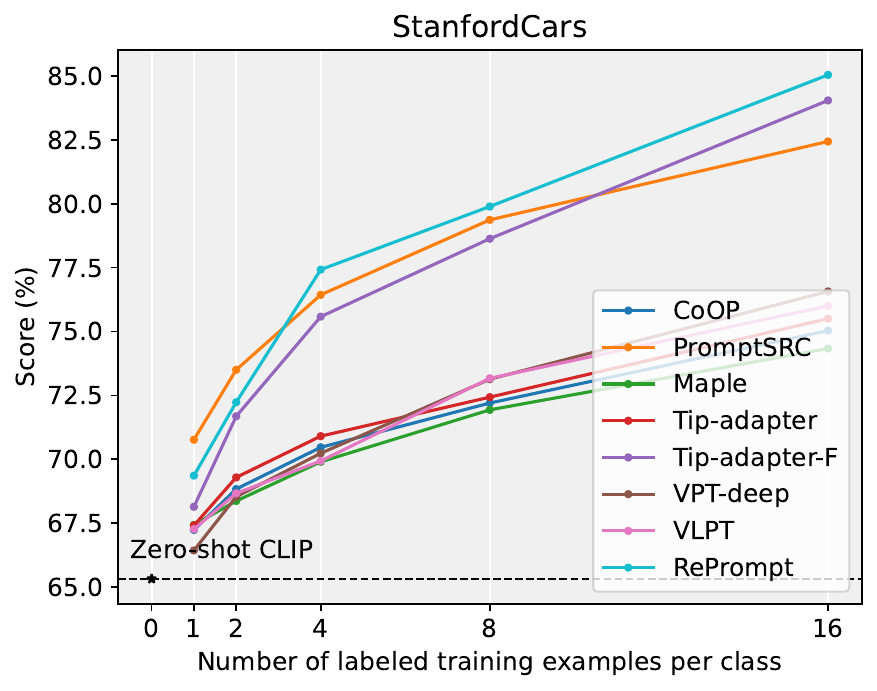}
	\end{minipage}
	\begin{minipage}{0.32\linewidth}
		\centering
		\includegraphics[width=0.9\linewidth]{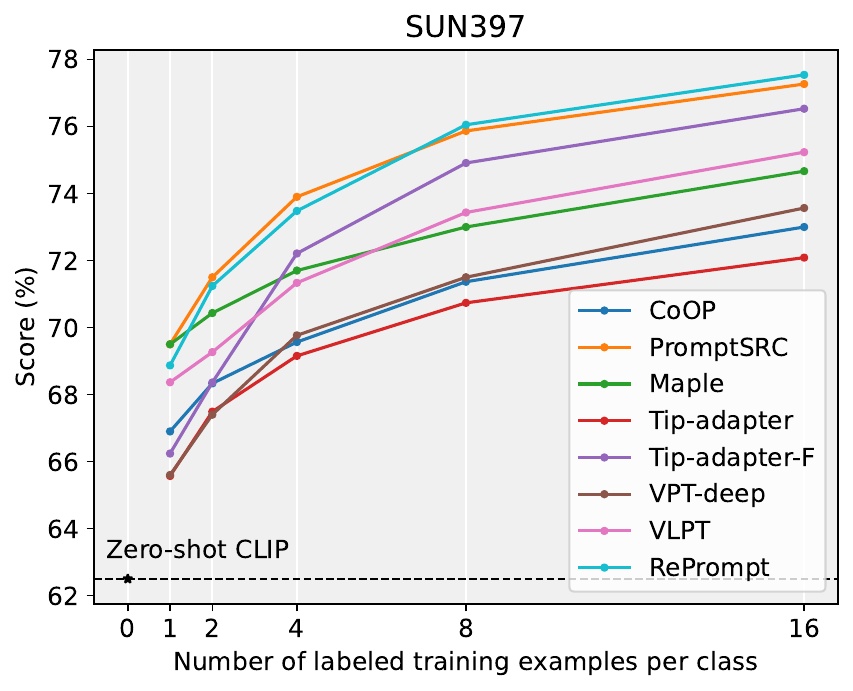}
	\end{minipage}
	\begin{minipage}{0.32\linewidth}
		\centering
		\includegraphics[width=0.9\linewidth]{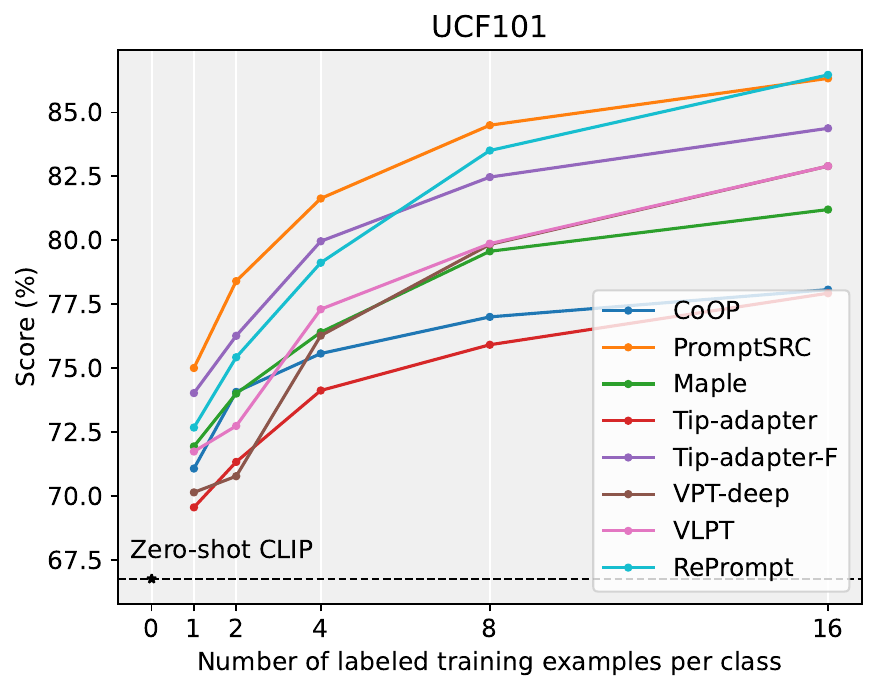}
	\end{minipage}
    \caption{\textbf{Main results over 11 datasets under the few-shot settings.} We report the average accuracy($\%$) of three runs for 1,2,4,8,16 shots. The proposed RePrompt achieves significant performance improvements on most downstream recognition datasets.}
   \label{fig:main_curves}
   \vspace{-3mm}
\end{figure*}

\subsection{Few-shot Classification}\label{subsec:fslearning}
\noindent\textbf{Datasets.} The RePrompt model is evaluated on 11 image classification datasets, covering various tasks such as object recognition (ImageNet1k~\cite{imagenet_paper} and Caltech101~\cite{Caltech101_paper}), fine-grained object recognition (Oxford Pets~\cite{oxford_pest_paper}, Flowers102~\cite{oxford_flowers_paper}, FGVC Aircraft~\cite{fgvc_aircraft_paper}, Stanford Cars~\cite{Stanford_cars_paper}, and Food101~\cite{food101_paper}), sense recognition (SUN397~\cite{sun397_paper}), remote sensing recognition (EuroSAT~\cite{eurosat_paper}), texture recognition (DTD~\cite{DTD_paper}), and action recognition (UCF101~\cite{ucf101_paper}).
we construct 1,2,4
,8,16-shot training sets with full categories and the entire test set.

\noindent\textbf{Baseline.}
We compare our proposed RePrompt with these existing prompt-based approaches:
CoOP~\cite{CoOP}, 
VPT-Deep~\cite{VPT}, 
Maple~\cite{MaPLe},
PromptSRC~\cite{promptsrc},
Tip-Adapter~\cite{tip-adapter},
Tip-Adapter-F~\cite{tip-adapter},
and VLPT.
(1) CoOP learns the context prompt concatenated with $\left[ \mathrm{CLASS} \right] $ as the input of text encoder. 
(2) VPT-Deep incorporates learnable visual prompts into each transformer layer of the visual encoder.
(3) VLPT jointly optimizes prompts across different modality encoders, following the methodologies of CoOP and VPT-Deep.
(4) Tip-Adapter constructs a cache-model-based adapter using few-shot training data. It is a model that performs few-shot predictions by using those retrieved samples in the end, simply by making some prototypes.
(5) Tip-Adapter-F is the variant where the adapter is fine-tuned.
(6) Maple learns prompts for the text and image branches simultaneously, rather than a separate side.
(7) PromptSRC uses the clip's original features to regularize the prompt learning.

\noindent\textbf{Training details.}
Prompt tuning, originally introduced in NLP, is extended to unified visual and language models by implementing it on a ViT model, specifically ViT-B/16, with 12 transformer layers similar to the text encoder.
The hyper-parameters of the retrieval modules are set as follows: $k_\mathrm{re}$ is set to 7, and 9 random prompts are initialized.
The first 7 layers of the vision transformer are equipped with retrieval-enhanced prompts.
The prompt and adapter are fine-tuned using the AdamW optimizer with a learning rate of 1e-3 and an epsilon value of 1e-4.
We adhere to the data preprocessing protocol of CoOP and freeze the parameters inherited from the pre-trained model during training.

\noindent\textbf{Results.} 
The performance of baseline approaches and our proposed RePrompt for few-shot image classification is presented in Fig.~\ref{fig:main_curves}. 
In general, parameter-efficient finetuning methods perform better than zero-shot classification (zero-shot CLIP) in scenarios with an average of over 11 datasets. RePrompt overall provides consistent improvements on most shots in comparison with all existing methods.
RePrompt demonstrates significant performance improvement, particularly with $+5.87\%$ accuracy on DTD and $+9.04\%$ accuracy on Stanford Cars, based on VLPT.
RePrompt substantially enhances performance on challenging datasets that contain a wide range of categories, such as ImageNet with 1000 classes and SUN397 with 397 classes.
We also observe that Reprompt achieves fewer improvements on Oxford Pets and Food101. 
This could be attributed to the noisy data and unique data distribution patterns in these datasets~\cite{CoOP,food101_paper,UPT}.

\noindent\textbf{Cross-dataset Evaluation.} 
We evaluate the cross-dataset generalization ability of RePrompt by learning on the 16-shot ImageNet setting and then transferring it directly to the remaining 10 datasets. 
Since this evaluation does not provide training images for other datasets, we remove the retrieval-based adapter module from this test. 
Table \ref{tab:cross_dataset} shows the performance comparison between CoOP, CoCoOP, Maple, and PrompSRC. On the ImageNet source dataset, RePrompt achieves the best performance and demonstrates a strong generalization performance by surpassing recent SOTA methods in 5 datasets. 
The lower performance on FGVC Aircraft, DTD, and UCF101 can be attributed to the different nature of their tasks, which require more specialized feature extraction. However,  we adopt the same retrieval database sampled from ImageNet.

\subsection{Domain Generalization}\label{subsec:domain}
Pre-trained vision-language models exhibit a robust ability for domain generalization ability~\cite{CoOP}. We assess the robustness of the proposed RePrompt model on out-of-distribution (OOD) datasets.

\noindent\textbf{Datasets.}
We follow CoOp~\cite{CoOP} and employ five datasets, namely, ImageNet, ImageNet V2~\cite{imagenetv2_paper}, ImageNet-Sketch~\cite{imagenet-sketch_paper}, ImageNet-A~\cite{imagenet-a_paper} and ImageNet-R~\cite{imagenet-r_paper}), to evaluate the generalization ability of RePrompt for out-of-distribution (OOD) data. 
Following the protocol, we train the model on the ImageNet dataset (as the source dataset) using a 16-shot setting and evaluate its performance on other domain-shifted datasets (as the target datasets). 
Consequently, we utilize the retrieval database from the ImageNet 16-shot experiment as the retrieval database for the target datasets.

\noindent\textbf{Results.} Table~\ref{tab:doman_generalization} summarizes the OOD experiment results, including the accuracy on both source datasets and target datasets.
RePrompt achieves optimal results on ImageNet V2 and ImageNet-Sketch while demonstrating performance comparable to PromptSRC~\cite{promptsrc} on ImageNet-R. 
These results indicate that Reprompt has better generalization for datasets with domain shifts.
RePrompt struggles on ImageNet-A and ImageNet-R. This is likely due to a lack of diversity in the retrieval database, only ImageNet, which limits generalization to adversarial and artistic images. In practical scenarios, we can improve the deficiency by expanding the search base and improving the search quality.

\begin{table*}
    \centering
    \caption{Comparison of RePrompt with existing advanced approaches on cross-dataset benchmark evaluation. The source model (ViT-B/16) is trained on ImageNet\cite{imagenet_paper}. 
    RePrompt achieves the best results on 5 of 10 datasets. The \underline{underline} represents the second best. 
    }
    \label{tab:cross_dataset}
    \vspace{-0.2cm}
    \resizebox{0.97\linewidth}{!}
    {
        \begin{tabular}{lcccccccccccc}
        \hline
        \multirow{3}{*}{Method}   & Source     & \multicolumn{10}{c}{Target} & \multirow{3}{*}{Avg.} \\
         & \multirow{2}*{ImageNet} & Caltech & Oxford & Standford & Flowers & \multirow{2}*{Food101} & FGVC & \multirow{2}*{SUN397} & \multirow{2}*{DTD} & Euro & \multirow{2}*{UCF101} \\
        ~   &          & 101     & Pets   & Cars      & 102     & ~  & Aircraft & ~ & ~     & SAT  & ~ & ~    \\
        \hline\noalign{\smallskip}
        CoOP\cite{CoOP} & \underline{71.51}  & 93.70 & 89.14 & 64.51 & 68.71 & 85.30 & 18.47 & 64.15 & 41.92 & \underline{46.39} & 66.55 & 63.88 \\
        CoCoOP\cite{CoCoOP} & 71.02  & \underline{94.43} & 90.14 & 65.32 & 71.88 & 86.06 & 22.94 & \underline{67.36} & 45.73 & 45.37 & 68.21 & 65.74 \\
        Maple\cite{MaPLe} & 70.72 & 93.53 & \textbf{90.49} & 65.57 & \underline{72.23} & \textbf{86.20} & \textbf{24.74} & 67.01 & \underline{46.49} & \textbf{48.06} & 68.69 & \textbf{66.30} \\
        PromptSRC\cite{promptsrc} & 71.27   & 93.60 & \underline{90.25} & \underline{65.70} & 70.25 & \underline{86.15} & 23.90 & 67.10 & \textbf{46.87} & 45.50 & \underline{68.75} & 65.81 \\
        \hline\noalign{\smallskip}
        \rowcolor{orange!5} RePrompt &\textbf{74.53}   &\textbf{94.65}  &90.16  &\textbf{67.76}  &\textbf{72.47}  &86.00  &\underline{24.33} &\textbf{67.71} &44.39 &41.74 &\textbf{69.28} &\underline{65.85}
 \\
        \hline
        \end{tabular}
    }
    \vspace{-3mm}
\end{table*}

\begin{table}
  \centering
  \caption{Main results under the domain generalization setting. We report the average accuracy ($\%$) of 16 shots over three runs.} \label{tab:doman_generalization}
  \vspace{-0.2cm}
  \scalebox{0.9}{
  \begin{tabular}{@{}l|ccccccc@{}}
   \hline
    Methods &ImagNet &V2\cite{imagenetv2_paper} &S\cite{imagenet-sketch_paper} &A\cite{imagenet-a_paper} &R\cite{imagenet-r_paper}  \\
    \hline
     CoOP\cite{CoOP} &71.51 &64.20 &47.99 &49.71  &75.21   \\
     CoCoOP\cite{CoCoOP} &71.02 &64.07 &48.75 &50.63  &76.18   \\
     UPT\cite{UPT} &72.63 &64.35 &48.66 &50.66  &76.24   \\
     Maple\cite{MaPLe} & 70.72 &64.07 &49.15 &\textbf{50.90}  &76.98 \\
    PromptSRC\cite{promptsrc} &71.27 &64.07 &49.55 &50.90 &\textbf{77.80} \\  
     \rowcolor{orange!5} RePrompt&\textbf{74.53} &\textbf{66.66} &\textbf{49.56} &49.77 &77.48  \\
    \hline
  \end{tabular}
  }
  \vspace{-3mm}
\end{table}

\begin{table}
\centering
\caption{
The main results on MVImgNet are under the few-shot multi-view settings.RePrompt achieved significant performance improvements due to its retrieval-enhanced module.
The methods marked with ``*'' are reproduced by us.
}
\label{tab:MVImgNet_few_shot}
\vspace{-0.2cm}
\scalebox{0.9}{
\begin{tabular}{l|cccccc}
  \hline
  \multirow{2}{*}{Model}  &\multicolumn{6}{c}{MVImgNet} \\
  & $1$-shot &$2$-shot &$4$-shot &$8$-shot  &$16$-shot &Ave.  \\
  \hline
Vanilla CLIP\cite{CLIP} &51.4  &51.4  &51.4 &51.4	&51.4  &51.4\\
CoOP* \cite{CoOP} &62.5 &67.8 &75.3 &79.8 &84.0 &73.9 \\
VLPT  &65.2  &69.7   &76.8	 &80.3   &84.9  &75.3\\
\rowcolor{orange!5}RePrompt    &\textbf{65.9}   &\textbf{73.6} 	&\textbf{81.1}	&\textbf{85.7}	&\textbf{90.8}  &\textbf{79.4}\\
    \hline                          
\end{tabular}
}
\vspace{-0.4cm}
\end{table}

\begin{table*}
\centering
\caption{We compare RePrompt with approaches that explicitly adapt CLIP. ViFi-CLIP only adds an average pooling layer on top of the final features to obtain video-level embeddings. The \underline{underline} represents the second best.}
\label{tab:video_understanding_few_shot}
\vspace{-0.2cm}
\scalebox{0.93}{
\begin{tabular}{l|ccccc|ccccc|ccccc}
  \hline
  \multirow{2}{*}{Model}  &\multicolumn{5}{c}{HMDB-51} & \multicolumn{5}{c}{UCF-101} & \multicolumn{5}{c}{SSV2}\\
  & $2$-shot & $4$-shot & $8$-shot  & $16$-shot &Ave. & $2$-shot & $4$-shot & $8$-shot  & $16$-shot &Ave. &$2$-shot & $4$-shot & $8$-shot  & $16$-shot &Ave.\\
  \hline
Vanilla CLIP\cite{CLIP} & 41.9	&41.9	&41.9	&41.9	&41.9 
&63.6	&63.6	&63.6	&63.6	&63.6 
&2.7	&2.7	&2.7	&2.7 &2.7\\
ActionCLIP\cite{wang2021actionclip}  & 54.3	&56.2	&59.3	&66.1	& 59.0
&76.7	&80.4	&87.6	&91.8	& 84.1
&4.8	&6.9	&9.1	&12.3  &8.3 \\
XCLIP\cite{ni2022expanding}   &60.5   &66.8  &69.3   &\textbf{71.7} &67.1
&89.0 &91.4 &94.7 &\textbf{96.3} &92.9
&6.6	&7.8  &\underline{9.9}	&\underline{13.7} &9.5 \\
A5\cite{ju2022prompting}      &46.7   &50.4  &61.3   &65.8 &56.1
&76.3 &84.4 &90.7 &93.0 &86.1
&4.5 &6.7 &7.2	&9.5 &7.0\\
ViFi-CLIP*\cite{rasheed2023fine}  &61.3 &65.3 &68.1 &70.1 &66.4
&\underline{90.4} &\underline{92.9} &\underline{94.4} &\underline{95.7} & \underline{93.3}
&7.2	&8.1 &\textbf{10.2}  &\textbf{13.9} &\textbf{9.7}\\
\rowcolor{orange!5} RePrompt       & \textbf{63.4}	&\textbf{67.6}	&\textbf{69.2}	&71.1	&\textbf{67.8} 
&\textbf{91.2}	&\textbf{93.5}	&\textbf{95.0}	&95.4	&\textbf{93.7} 
&\textbf{7.6}	&\textbf{9.3}	&9.4	&12.1  &\underline{9.6}\\ 
    \hline                          
\end{tabular}
}
\vspace{-0.3cm}
\end{table*}

\subsection{Additional Few-shot Classification}\label{subsec:other_classification}
We investigate the ability of RePrompt to bridge the modality gap in video and multi-view domains. 

\noindent\textbf{Datasets.}
MVImgNet~\cite{MVImgNet} is a 3D generic dataset that comprises multi-view images, which is a soft bridge between 2D and 3D. 
It contains 6.5 million frames from 219188 videos, accompanied by comprehensive annotations. These frames covers real-life objects across from 238 classes.
Nevertheless, MVImgNet displays a long-tailed distribution of data. Consequently, we eliminate categories with fewer than 16 samples and samples with fewer than 28 frames.
In the end, we obtained a subset of MVImgNet, consisting of a total of 220 classes. 
We adopt the same few-shot setting as used in few-shot image classification.
The video understanding task covers HMDB-51~\cite{kuehne2011hmdb}, UCF-101~\cite{ucf101_paper} and SSv2~\cite{goyal2017ssv2}, where each dataset consists of 2/4/8/16-shot training sets with full categories. The model are evaluated on the first validation split for HMDB-51, UCF-101, and the whole validation split for SSV2.

\noindent\textbf{Training details.}
We adopt a sparse frame-sampling strategy~\cite{wang2016tsn} and configure the number of frames as 16 for multi-view classification and 32 for video understanding. For the video understanding task, models are pre-trained on Kinetics-400~\cite{k400} to bridge the modality gap and follow a single-view inference~\cite{rasheed2023fine}.
he retrieval setup aligns with that of few-shot image classification.
However, the training hyperparameters remain consistent with those of ViFi-CLIP-Prompting~\cite{rasheed2023fine}.
The minimum retrieval unit in the database is a frame-level embedding, aligned with the input data. The weights of the ``Rb adapter" are sourced from another database, where the minimum retrieval unit is a video-level embedding.
The use of pooling for simple frame-level temporal aggregation enables the exchange of temporal and multi-view cues in CLIP representations.

\noindent\textbf{Results.}  
Table~\ref{tab:MVImgNet_few_shot} demonstrates the RePrompt consistently improves performance as the number of shots increases in multi-view classification. 
For example, it achieves a gain of $+4.2\%$ compared to VLPT. 
This improvement can be attributed to the retrieval process, which brings more reference information from different perspectives.
Table~\ref{tab:video_understanding_few_shot} illustrates the effectiveness of RePrompt compared to other CLIP-based approaches for videos. Notably, RePrompt achieves larger gains in low-shot data scenarios, where models are more susceptible to overfitting.

\begin{figure}[h]
  \centering
  \vspace{-0.2cm}
  \includegraphics[width=0.9\linewidth]{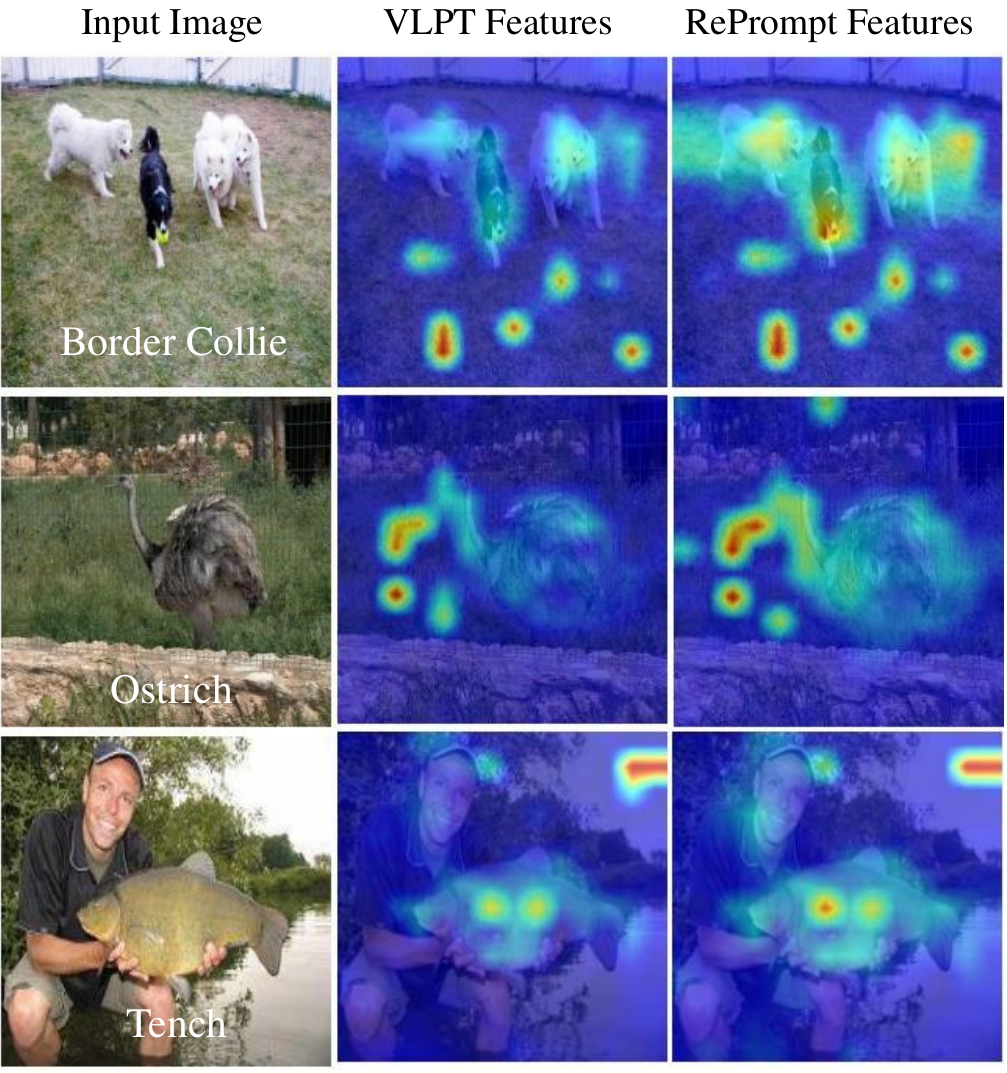}
  \caption{Visualization of attention response map between retrieval-enhanced visual prompts and image patch tokens. The mean self-attention map is from the last vision transformer layers. 
  }
  \label{fig:attention_map}
  \vspace{-5mm}
\end{figure}

\subsection{Ablation Study}\label{subsec:ablation}

\begin{table*}
  \centering
  \caption{Component ablation studies over 11 datasets with 16-shot setting. The average accuracy of RePrompt has been steadily improved through the gradual introduction of retrieval enhancement modules.
  We reproduce methods marked with ``*'' on current datasets. ``Reprompt+Synthetic data'' means introducing synthetic data on the basis of RePrompt.
  }
  \label{tab:componet}
  \vspace{-0.2cm}
  \scalebox{1.1}{
  \begin{tabular}{@{}l|cccccccccccc@{}}
    \hline
    Method & \rotatebox{90}{EuroSAT} 	& \rotatebox{90}{Caltech101}  & \rotatebox{90}{Flowers102} 	& \rotatebox{90}{Food101} &	\rotatebox{90}{FGVCAircraft} 	&\rotatebox{90}{DTD} 	&\rotatebox{90}{OxfordPets}	 &\rotatebox{90}{StanfordCars} 	&\rotatebox{90}{UCF101}	&\rotatebox{90}{SUN397} 	&\rotatebox{90}{ImageNet}  &\rotatebox{90}{average} \\
    \hline
    CoOP*\cite{CoOP} &80.87 	&95.50 	&92.20 	&86.30 	&32.57 	&67.17 	&92.87 	&75.03 	&78.07 	&73.00 	&71.20 	&76.80 \\
    VPT-deep*\cite{VPT} &85.83 	&95.63 	&95.00 	&86.30 	&41.53 	&68.93 	&93.20 	&76.57 	&82.90 	&73.57 	&70.40 &79.80 \\
    PromptSRC* &92.30 &96.03  &97.63  &87.47  &48.03  &73.27  &93.73  &82.43  &86.33  &77.27  &72.93 &82.49 \\
    Maple*\cite{MaPLe} &86.20 &95.20 &93.57 &87.47 &37.47 &67.53 &93.77 &74.33 &81.20 &74.67 &71.97 &78.49\\
    \hline
    Tip-Adapter\cite{tip-adapter} &78.09  &95.09 &94.62 &86.52 &39.68 &65.90 &91.98 &75.50 &77.93 &72.09 &70.75 &77.10 \\
    Tip-Adapter-F*\cite{tip-adapter} &88.74 	&95.94 	&96.79 	&87.28 	&46.32 	&72.87 &92.89 	&84.04 	&84.38 	&76.53 	&73.32 	&81.74 \\
    \hline
    +VLPT  & 89.50 	& 95.57 	& 95.37 	& 87.17 	& 48.04	& 67.83 	& 93.60 	& 76.00 	& 82.90 	& 74.20 	& 71.42 	& 80.14 \\
    +Rg training loss & 92.48	& 96.55 	& 94.72 	&\textbf{87.56} 	& 47.94	& 69.74 	& 93.54 	& 74.18 	& 83.53 	& 75.20 	& 72.00 	& 80.68 \\
     +Re visual prompt &92.60	&96.55	&96.79	&87.57	&48.27	&70.63	&93.62	&77.85	&83.95	&75.60	&72.20	&81.42 \\
   \rowcolor{orange!5} +Rb adapter(RePrompt) &\textbf{92.91}	&96.51	&97.16	&87.42	&50.32	&73.70	&93.76	&85.04	&86.47	&\textbf{77.54}	&\textbf{74.53}	&83.21 \\
\hline
CaFO-F*\cite{zhang2023prompt}  &91.72 	&\textbf{97.28} 	&97.97 	&87.15 	&54.82 	&71.34  &93.89 	&83.85 	&86.10 	&76.62 	&74.48 	&83.20\\
Reprompt+Synthetic data &91.95 &96.80 &\textbf{98.29} &87.23 &\textbf{55.21} &\textbf{76.18} &\textbf{93.92} &\textbf{87.55} &\textbf{87.34} &77.39 &74.23 &\textbf{84.19} \\
    \hline
  \end{tabular}
  }
  \vspace{-0.3cm}
\end{table*}

\begin{figure*}[h]
  \centering
  \includegraphics[width=1.0\linewidth]{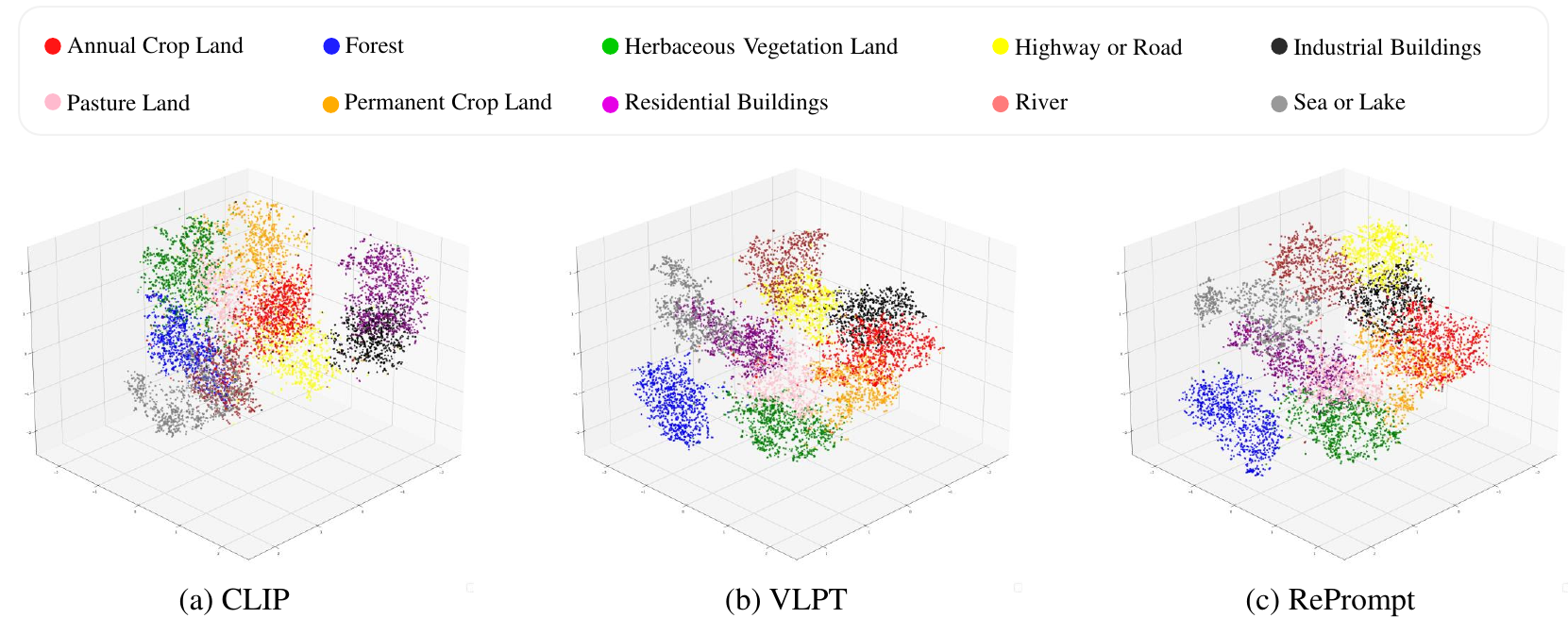}
  \caption{Visualization of different learned visual feature manifolds via t-SNE. From left to right, we have CLIP, VLPT, and RePrompt. The image features extracted by RePrompt discriminate more along the z-axis.
  }
  \label{fig:t-SNE-3}
  \vspace{-0.5cm}
\end{figure*}

\noindent\textbf{Component ablation.} We investigate the effectiveness of RePrompt and report results in Table~\ref{tab:componet}. 
A progressive approach of incorporating retrieval modules is employed to demonstrate the efficacy of enhancement measures at various stages. The average accuracy of RePrompt consistently improves as we incrementally introduce retrieval augmentation: Retrieval-guided training denoted as $+$Rg training loss, Retrieval-enhanced visual prompt denoted as $+$Re visual prompt, and Retrieval-based adapter denoted as $+$Rb adapter.
We also evaluated the Tip-adapter, which solely utilizes the retrieved samples as prototypes. The findings indicate that retrieval-enhanced visual prompts significantly enhance retrieval-based classification.

Furthermore, we visualize the attention map in the last layer of the image encoder for both VLPT and RePrompt to illustrate the impact of retrieval-enhanced visual prompts. The visualization results, displayed in Fig.~\ref{fig:attention_map}, reveal that RePrompt has more pronounced self-attention responses, characterized by expanded areas of interest and heightened attention values.
This enhancement correlates with a performance improvement of "+1.28$\%$" since the inclusion of "+Re visual prompt".

The noticeable improvement achieved with the "+Rb adapter" can be ascribed to its capability to introduce additional degrees of freedom for the query features, which is unattainable with the conventional Tip-Adapter-F.
Tip-Adapter operates as a training-free approach, leveraging the similarity of image features to aid in classification. Despite attempts to mitigate this limitation through fine-tuning in Tip-Adapter-F, the query features remain unchanged.
The limited ability of Tip-Adapter-F to manipulate query features significantly hampers its performance.

To further substantiate these observations, we employ t-SNE~\cite{tSNE} to visualize the 3D manifold of test features extracted from the CLIP, VLPT, and RePrompt models on the EuroSAT dataset. The t-SNE visualization results, presented in Fig.~\ref{fig:t-SNE-3}, clearly depict that in the high-dimensional classification space, RePrompt achieves a much more distinct separation of image features, with feature distributions more aligned along the z-axis, distinguishing between various categories. Conversely, Tip-Adapter-F, constrained by a frozen visual encoder, exhibits t-SNE visualization results that closely resemble those of CLIP, with considerable overlap in feature distribution. This overlap elucidates why Tip-Adapter performs suboptimally on remote sensing data such as EuroSAT.

\noindent\textbf{Expanding retrieval database.}
We follow recent approaches~\cite{udandarao2022sus, zhang2023prompt} that leverage a powerful text-to-image generative model, Stable Diffusion~\cite{rombach2022high}, to construct a pseudo training set or expand the existing training set. The $\Acute{K}$ is set to 2.
In the bottom row of Table \ref{tab:componet}, our results show that RePrompt benefits from synthesis data and achieves state-of-the-art performance (average accuracy 84.19$\%$). 
Compared to CaFO~\cite{zhang2023prompt}, an improvement method of Tip-Adapter-F, RePrompt achieves an average precision improvement of +0.99 $\%$.
We can observe that while synthetic data generally boosts performance across most datasets, the performance on EuroSAT drops slightly. The performance drop in EuroSAT after introducing synthetic data could be due to the synthetic data not being well-aligned with the characteristics of satellite images. In real scenarios, we can improve this deficiency by introducing more real satellite images.


\noindent\textbf{Discussions on ReConv and retrieved samples.}\label{subsec:Reprompt_details}
We explore the impact of the network structure, which is responsible for integrating tokens and queries, on the performance outcomes of the ReConv model. 
To this end, we conduct a supplementary experiment on ImageNet, where we substitute the convolution layer in ReConv with different layers. 
ReRNN is a variant in which the convolution layer is replaced with an LSTM layer~\cite{LSTM, RNN}. 
ReMLP replaces the convolution layer with a multi-layer perceptron layer. 
Additionally, the quality of high-retrieval samples also contributes to performance; hence, we examine the effect of utilizing original features and random sample features as inputs. 
ReConv($z_q$) is a variant that exclusively uses $z_q$ as input.
The ReConv+Random employs random sample features from the retrieval database.
The experimental setup for these variants aligns with that of ReConv, maintaining a comparable number of parameters across both models.

As demonstrated in Table~\ref{tab:reconv_var}, ReConv marginally outperforms the other variants across all shot levels. 
\textbf{These results suggest that the specific network structure used for token and query fusion may not significantly affect the overall model performance.}
Instead, the choice of network structure should take into account the computational cost and optimization challenges, particularly as task complexity increases. 
However, the quality of retrieval samples plays a crucial role, implying that expanding the search library and enhancing the search methodologies are of greater importance.

\begin{table}[h]
 \vspace{-2mm}
  \centering
  \caption{Ablation study of the visual prompt learner across different input strategies and network structures. The first two rows showcase results using RNN and MLP structures, respectively. 
  The middle two rows illustrate the impact of using original image features and random sample features as inputs. 
  ReConv consistently outperforms the other variations, demonstrating its effectiveness across varying shot scenarios.
  }
  \label{tab:reconv_var}
  \vspace{-2mm}
  \scalebox{1.0}{
  \begin{tabular}{@{}l|ccccc@{}}
    \hline
    method  & 16-shot	& 8-shot	& 4-shot	& 2-shot	& 1-shot
 \\
    \hline
    ReMLP  &74.21 &73.02 &71.40 &70.33 &69.50 \\
    ReRNN  &74.27 &72.60 &71.12 &70.32 &69.49 \\
    \hline
    ReConv($z_q$)  &74.34 &72.51 &70.84 &70.42 &69.33 \\
    ReConv+Random &74.03 &72.43 &70.68 &69.98 &66.96 \\
    \hline
    \rowcolor{orange!5} ReConv &\textbf{74.53} &\textbf{73.36} &\textbf{71.84} &\textbf{70.68} &\textbf{69.87}  \\
   \hline
  \end{tabular}
  }
   \vspace{-2mm}
\end{table}

\begin{table}[h]
  \centering
  \caption{Comparison of training efficiency for different methods on 16-shot ImageNet. All experiments are trained with batch 16 on one RTX309 GPU.}
  \vspace{-0.2cm}
  \label{tab:parameter and training time}
  \setlength{\tabcolsep}{1pt}
  \scalebox{1.1}{
  \begin{tabular}{@{}l|ccccccc@{}}
   \hline
    Methods &Acc.($\%$) &Param.(M) &Train. &Epochs\\
    \hline
     CoOP\cite{CoOP} &71.20 &0.4 &14h40min &200  \\
     Tip-Adapter-F\cite{tip-adapter} &73.32 &8.19 &5min &20 \\
     VLPT &71.42 &0.45 &15h50min &100 \\
    \hline
     Wo Retrieval &71.91 &11.11 &3h30min &20 \\
     RePrompt &74.53 &11.11 &4h  &20 \\
    \hline
  \end{tabular}
  }
  \vspace{-2mm}
\end{table}

\begin{table}[h]
  \centering
  \caption{Comparison of inference efficiency for different methods on 16-shot ImageNet. All experiments are trained with batch 16 and tested with batch size 32 on one RTX309 GPU.}
  \label{tab:inference time}
  \vspace{-2mm}
  \setlength{\tabcolsep}{1pt}
  \scalebox{1.1}{
  \begin{tabular}{@{}l|ccc@{}}
   \hline
    Methods &Acc.($\%$)  &Inference time (ms) &GFLOPs(inference)\\
    \hline
     CoOP*\cite{CoOP} &71.20  &299.64 &162.5 \\
     Tip-Adapter-F*\cite{tip-adapter} &73.32  &10.5 &42.5\\
     VLPT &71.42 &214.84 &162.5\\
    \hline
     Wo Retrieval &71.91  &17.57 &76.5 \\
     RePrompt &74.53 &57.60 &76.5 \\
    \hline
  \end{tabular}
  }
  \vspace{-2mm}
\end{table}

\noindent\textbf{Training and inference compute cost analysis.}
The compute cost and efficiency analysis is listed in Table~\ref{tab:parameter and training time} and Table~\ref{tab:inference time} for 16-shot classification on ImageNet. 
``Wo Retrieval" is a comparative model with the same number of learnable parameters as RePrompt. 
RePrompt shows a promising balance between accuracy and inference efficiency, achieving 74.53$\%$ accuracy with relatively low inference time (57.60 ms) and moderate computational demand (76.5 GFLOPs). 
Furthermore, we reduce the training time to only 20 epochs, in contrast to CoOP.
\textbf{In comparison to Tip-Adapter-F, RePrompt provides undeniable performance advantages and superior cross-domain adaptability.} 
Additionally, the retrieval time during inference is about 40 ms, which is deemed acceptable for few-shot image classification. 

Overall, these results highlight the trade-offs between training duration, model complexity, computational efficiency, and performance, emphasizing the effectiveness of RePrompt in leveraging retrieval for improved few-shot classification on high-complexity tasks.

\section{Retrieval Discussions}\label{sec:more retrieval ablation}
\begin{figure}[h]
\centering
\includegraphics[width=0.8\linewidth]{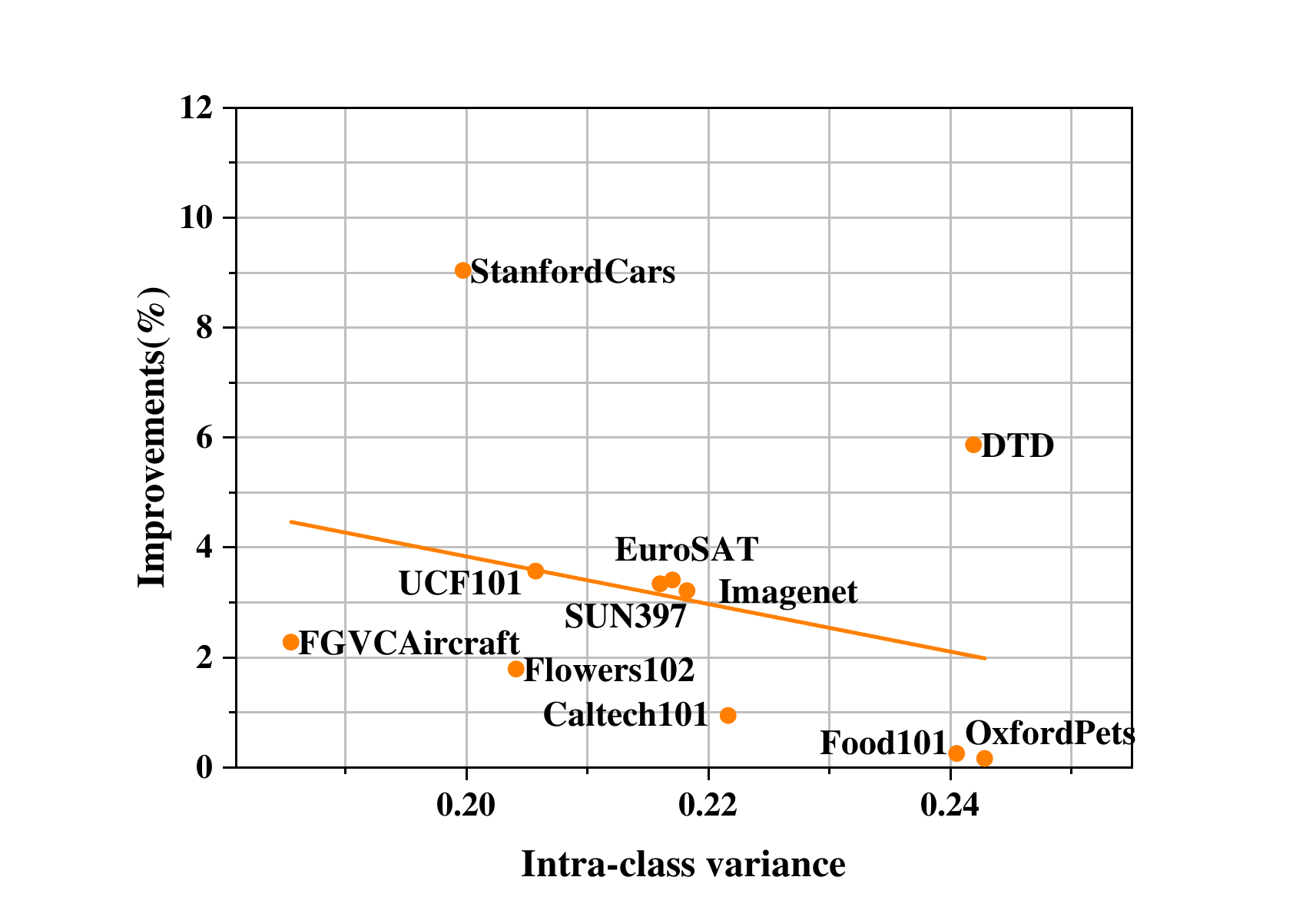}
\vspace{-2mm}
\caption{\textbf{Visualization of the Correlation Between Performance Improvements and Intra-class Visual Variance.} The graph demonstrates a general trend where the performance improvements offered by RePrompt tend to decrease as the intra-class visual variance increases. This suggests a sensitivity of the RePrompt's efficacy to the homogeneity of visual features within classes.}
\label{fig:correlation}
\vspace{-4mm}
\end{figure}

Given that the retrieval database is constructed from training data in a few-shot setting, the performance enhancements observed with RePrompt are intimately linked to data statistics. 
To delve deeper, we investigate the correlation between the efficacy of retrieval enhancement and the intra-class variance of visual features.
This specific relationship is depicted in Fig.~\ref{fig:correlation}, illustrating the association between intra-class visual variance and performance improvement. Aside from the outliers—``StanfordCars'' and ``DTD''—the remaining datasets exhibit a discernible pattern. Notably, the labels for StanfordCars correspond to vehicle codes, and those for DTD to texture descriptions, categories where CLIP's classification capabilities manifest notable deficiencies. Consequently, results from these two datasets are excluded from the general analysis to delineate retrieval enhancement patterns more accurately.

\begin{table}[h]
  \centering
  \vspace{-3mm}
  \caption{Ablation study on different visual prompt numbers over Fgvc Aircraft and Oxford Pets with few-shot settings. The retrieval-enhanced prompt is mainly affected by the convolution kernel size of REConv.}
  \label{tab:prompt_num_more}
  \vspace{-0.2cm}
  \resizebox{1.\columnwidth}{!}{
  \begin{tabular}{l|c|c|c|c|c|c|c}
    \hline
    $k_\mathrm{re}$+2 & Dataset & 16-shot	& 8-shot	& 4-shot	& 2-shot	& 1-shot  &Ave \\
    \hline
    2+2  & \multirow{3}{*}{FGVC Aircraft} &53.05	&45.84 &39.39 &34.86 &31.05  &40.838 \\
    7+2  & &54.61 &47.94 &40.05 &35.61 &\textbf{31.23} &41.888 \\
    14+2 & &\textbf{54.91} &\textbf{47.97} &\textbf{40.17} &\textbf{35.79} &31.17  &\textbf{42.002} \\
    \hline
    2+2  & \multirow{3}{*}{Oxford Pets} &94.58 &\textbf{93.84} &93.21 &\textbf{92.97} &91.55 &93.23 \\
    7+2  & &\textbf{94.85} &93.57 &\textbf{93.62} &92.94 &\textbf{91.82} &\textbf{93.36} \\
    14+2 & &94.47 &93.65 &93.16 &93.21 &91.41 &93.18 \\
    \hline
    2+2  & \multirow{3}{*}{ImageNet} &74.50	&73.23 &\textbf{71.51} &\textbf{70.53} &69.46  &71.85\\
    7+2  & &\textbf{74.53} &\textbf{73.28} &71.38 &70.40  &\textbf{69.87} &\textbf{71.89}\\
    14+2 & &74.40 &73.11 &71.29 &70.35 &69.64  &71.76\\
    \hline
  \end{tabular}}
  \vspace{-2mm}
\end{table}

\begin{table}[h]
\vspace{-0.2cm}
  \centering
  \caption{Ablation studies on retrieving $n$ samples for each class over Fgvc Aircraft and Oxford Pets with few-shot settings. The retrieval-guided training is weak in low-shot settings, since the model may require reference on the $k$-nn classifier.}
  \label{tab:loss_n_more}
  \vspace{-0.2cm}
  \resizebox{1.0\columnwidth}{!}{
  \begin{tabular}{l|c|c|c|c|c|c}
    \hline
    $\boldsymbol{n}$ & Dataset & 16-shot	& 8-shot	& 4-shot	& 2-shot	& 1-shot   \\
    \hline
    16 & \multirow{5}{*}{FGVC Aircraft} &54.40	&$-$	&$-$	&$-$	&$-$ \\
    8  & &54.61	&\textbf{47.94}	&$-$	&$-$	&$-$ \\
    4  & &\textbf{54.79}	&47.28	&\textbf{40.11}	&$-$	&$-$ \\
    2  & &54.04	&47.31	&39.96	&\textbf{35.81}	&$-$\\
    1  & &54.58	&47.31	&40.05	&35.43	&\textbf{31.23}\\
    \hline
    16 & \multirow{5}{*}{Oxford Pets} &94.60 	&$-$	&$-$	&$-$	&$-$ \\
    8  & &94.49 	&93.43 	&$-$	&$-$	&$-$ \\
    4  & &94.55 	&\textbf{93.57} 	&\textbf{93.62} 	&$-$	&$-$ \\
    2  & &94.69 	&93.51 	&93.21 	&\textbf{92.94} 	&$-$\\
    1  & &\textbf{94.85} 	&93.46 	&93.57 	&92.70 	&\textbf{91.82} \\
    \hline
    16  & \multirow{5}{*}{ImageNet-1k} &74.30	&$-$	&$-$	&$-$	&$-$\\
    8  	& &\textbf{74.53}	&73.11	&$-$	&$-$	&$-$\\
    4  	& &74.26	&\textbf{73.30}	&\textbf{71.52}	&$-$	&$-$\\
    2   & &74.31	&73.16	&71.50	&\textbf{70.40}	&$-$\\
    1   & &74.37	&73.23	&71.24	&70.38	&\textbf{70.02}\\
    \hline
  \end{tabular}}
  \vspace{-0.1cm}
\end{table}

\subsection{Analysis of Retrieval}\label{relation_anysls}
Experiments are conducted across FGVC Aircraft, Oxford Pets and ImageNet-1K to quantitatively assess the influence of retrieval in various few-shot scenarios. 
FGVC Aircraft has the smallest intra-class visual variance, while Oxford Pets has the largest; ImageNet-1K falls intermediate between these two.
By comparing the optimal retrieval parameter adjustments among these datasets, we can analyze the influence of intra-class visual variance on the retrieval mechanism. 

The increasing value of the \textbf{retrieval-enhanced prompt number $k_\mathrm{re}+2$} indicates that RePrompt benefits from additional retrieved information.
As illustrated in Table~\ref{tab:prompt_num_more}, the optimal $k_\mathrm{re}$ parameter varies among the three datasets. 
Specifically, we observe that $k_\mathrm{re}=14$ consistently underperforms across all few-shot settings for Oxford Pets.
In contrast, $k_\mathrm{re}=14$ demonstrates improved average performance on FGVC Aircraft, underscoring its utility in scenarios with lesser intra-class variance.
\textbf{This suggests that the retrieved information is detrimental and lacks confidence in contexts with significant intra-class visual variance.}

The factor $n$ determines \textbf{the number of retrieval samples, $\left| \boldsymbol{K} \right| = C \times \boldsymbol{n}$}, used in retrieval-guided training loss. 
Increasing the value of $n$ signifies a more relaxed constraint on retrieval. 
As shown in Table~\ref{tab:loss_n_more}, $\boldsymbol{n}=1$ achieves optimal performance in the 16-shot setting on Oxford Pets, which implies that RePrompt prefers the output of CLIP with retrieval-enhanced prompt. 
In low-shot scenarios(e.g., $\boldsymbol{n}=2/1$ under 2/1-shot setting ), RePrompt may require more additional references due to the scarcity of training data.


\textbf{Based on the above analysis, we summarize two empirical rules as follows:} 1) For a dataset with low or high intra-class visual variance, it is imperative to employ a retrieval model with correspondingly strong or weak capabilities to guarantee stable performance improvements across scenarios.
For example, if the dataset has a large intra-class visual variation, we need to control the impact of the research branch on the results by limiting the number of retrieved samples and retrieval-enhanced prompt.
2) In low-shot settings, the inference outcomes are dominantly influenced by the retrieval results.

\section{Conclusion}
In this paper, we propose RePrompt, a retrieval-based framework that enhances the performance of visual-language modes on few-shot classification tasks.
Specifically, we integrate the retrieved results from a retrieval cache model into model inference using two approaches: prompts and predicted distributions.  
Additionally, We introduce a prompt learner that dynamically adapts prompts based on retrieval results and an auxiliary classification adapter. 
To regulate the degree of retrieval enhancement, we introduce the prior distribution obtained from semi-parametric retrieval into the cross-entropy loss to guide prompt tuning.

Extensive experiment results prove that the proposed method achieves superior performance over other prompt-learning methods in few-shot learning and comparable results on domain generalization.  
More significantly, we extend the RePrompt approach from traditional image datasets to more challenging tasks, including video and multi-view image datasets.
Finally, we summarize the relationships between the properties of external memory and retrieval enhancement mechanisms. This conclusion stems from a quantitative analysis of the relationship between the hyperparameters of retrieval enhancement and the data distribution of external memory.

We hope that our findings inspire further research in these promising directions:
1) Extending prompt learning to additional downstream tasks, particularly dense prediction tasks such as semantic segmentation and object detection.
2) Investigating the application of retrieval techniques to address challenges, such as long-tail data.
Our study lays the groundwork for future research in enhancing visual-language models through retrieval mechanisms, and we encourage further investigations into these critical areas.

\bibliographystyle{IEEEtran}
\bibliography{reference}

\vfill

\end{document}